\documentclass[11pt]{article}

\usepackage{ACL2023}
\usepackage{graphicx}
\usepackage{subcaption}
\usepackage{enumitem}
\usepackage{graphicx}  % Required for including images
\usepackage{float} 
\usepackage{amssymb}% For controlling the float behavior (optional)`'
\usepackage{times}
\usepackage{latexsym}
\usepackage{amsmath}
\usepackage{multicol}
\usepackage{multirow}
\usepackage{graphicx}
\usepackage{booktabs}
\usepackage{afterpage}
\usepackage{colortbl}  % For coloring rows
\usepackage{xcolor}    % For custom colors
\usepackage[T1]{fontenc}
\usepackage[utf8]{inputenc}

\usepackage{microtype}

\usepackage{inconsolata}

 \title{Just \emph{KIDDIN'} : \textit{K}nowledge \textit{I}nfusion and \textit{D}istillation for \textit{D}etection of \textit{IN}decent Memes  
    \vspace{0.5em} % Adjust to move closer to the title if necessary
    \small
    {\color{red} \textbf{Warning: This paper contains content that may be potentially offensive or upsetting.}}
}

\author{
  Rahul Garg\thanks{\ \ Equal contribution.} \\ IIIT Hyderabad \\ Hyderabad, India \\ {\small \texttt{rahul.garg@research.iiit.ac.in}}
  \And
  Trilok Padhi\footnotemark[1] \\ Georgia State University \\ Atlanta, GA, USA \\ {\small \texttt{tpadhi1@student.gsu.edu}}
  \And
  Hemang Jain \\ IIIT Hyderabad \\ Hyderabad, India \\ {\small \texttt{hemang.jain@students.iiit.ac.in}}
  \AND
  Ugur Kursuncu \\ Georgia State University \\ Atlanta, GA, USA \\ {\small \texttt{ugur@gsu.edu}}
  \And
  Ponnurangam Kumaraguru \\ IIIT Hyderabad \\ Hyderabad, India \\ {\small \texttt{pk.guru@iiit.ac.in}}
}

\begin{document}
% \maketitle
\maketitle

% Try to incorporate seen and unseen into a single table. Can remove deltas. 
% Remove GAT + GCN Theory
% Should add all 4 rows of combinations of Hops and Encoder
% Introduction reduction of 1 para atleast.
% Eqn1 s_i (inconsistency of boldness) 301 and 264
% ConceptNet C and Caption C- Confusing
% Remove knowledge column
% Text over arrows + break the vector into two
% change name of consistency loss in figure
% Fix subscripting in the figure and explain px in caption
% Consistency loss arrow needs to be removed. It has to be a end result. 
% Change figure 2 to be a flow rather than this. 
% 51 english 
% caption can be reduced for figure 2
% Figure 3: Move the legend down and reduce its size
% Line 267, 264 and 271 - Captions - Ci and C is Conceptnet - confusing
% w_clip_llava fix
% hatememe hatefulmeme consistency
% Why roberta, why minilm
% ablation of graph types (Might not be possible)
% Pooling section can be removed 

\begin{abstract}
Detecting toxicity in online multimodal environments, such as memes, remains a challenging task due to the complex contextual connections across modalities (e.g., text and visual), which demand both common-sense reasoning and contextual awareness. To bridge this gap, we propose a hybrid neurosymbolic framework that unifies (1) distillation of implicit contextual knowledge (e.g., sarcasm, cultural references) from Large Vision-Language Models (LVLMs) and (2) infusion of explicit relational semantics through sub-graphs from Knowledge Graphs (KGs). Experimental results on two benchmark datasets show the superior performance of our approach, \emph{Knowledge-Infused Distilled Vision-Language Model (KID-VLM)}, over the state-of-the-art baselines across AUC and F1, with improvements of 0.5\%, and 10.6\%, respectively, in HatefulMemes Benchmark across variants. Further, KID-VLM demonstrates better generalizability and achieves the best performance across all baselines in the HarMeme Dataset with a 6.3\% and 3.2\% in F1 and AUC.
Given the contextual complexity of the toxicity detection, KID-VLM showcases the significance of learning compact models (\textasciitilde
500M parameters) from both explicit (i.e., KG) and implicit (i.e., LVLMs) contextual cues incorporated through a hybrid neurosymbolic approach. 
Our codes and pretrained models are publicly available\footnote{\url{https://github.com/SWAN-AI/Knowledge-Infused-Distilled-VLMs}}.
% KID-VLM provides a scalable approach to recognize toxic content for safer online environments in real-world applications. 

% Github\footnote{\url{https://github.com/SWAN-AI/Knowledge-Infused-Distilled-VLMs}}.

%Furthermore, the proposed framework can be easily extended to applications beyond toxicity detection, such as misinformation detection, sarcasm detection or any task that requires a deep understanding of multimodal content with context-aware reasoning.
\end{abstract}

\section{Introduction}
\label{sec:intro}
\begin{figure}[hbt]
    \begin{minipage}{0.5\textwidth}
        \centering
        \includegraphics[height=8cm, width=8cm]{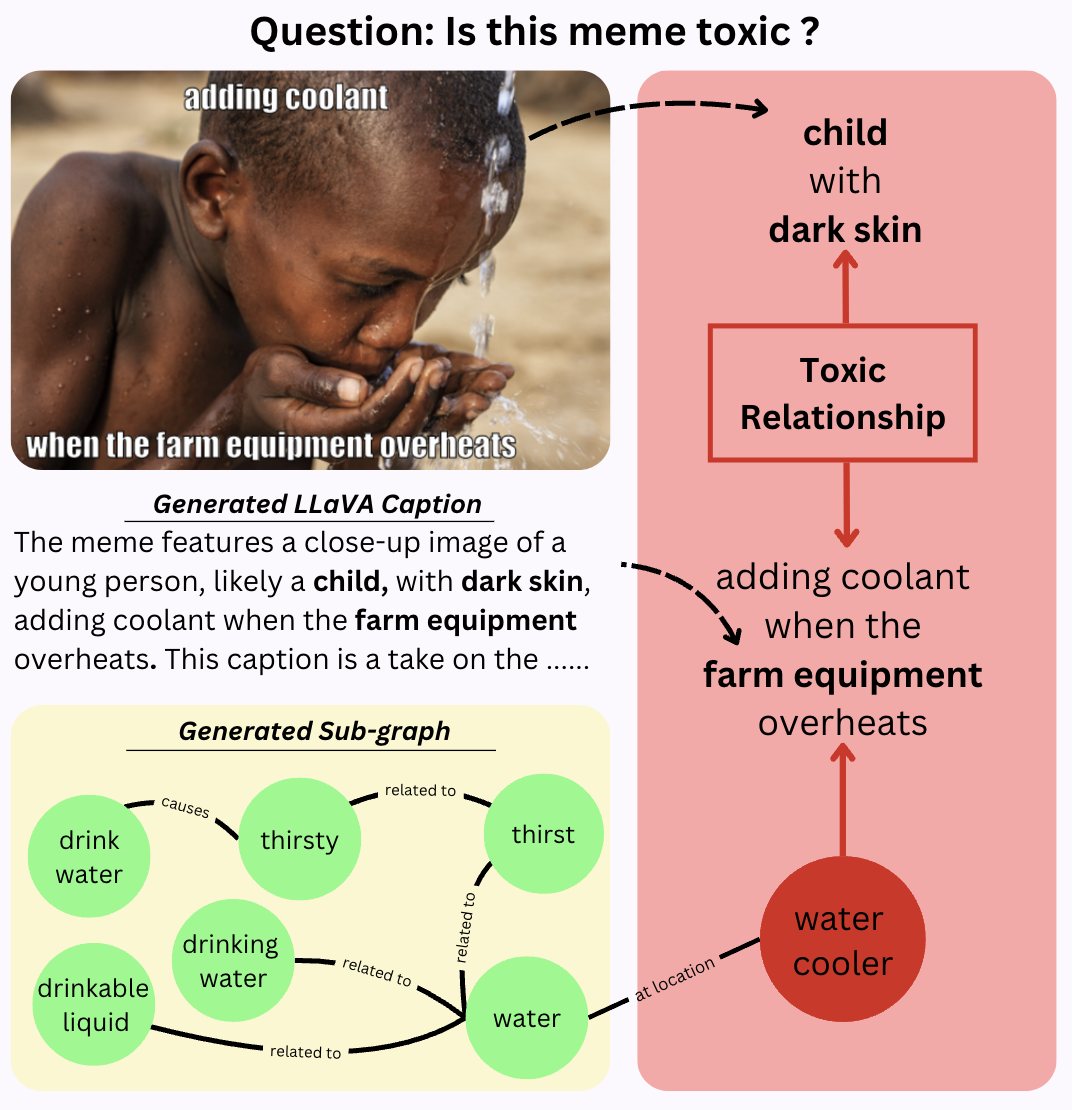}
        \caption{Given a meme, we aim to derive the answer by joint reasoning over the knowledge from LVLM, the KG (green box), and reason over toxicity (red box).
}
\vspace{-20pt}  % Adjust as needed to reduce space
        \label{fig:kid_vlm_motivation}
    \end{minipage}
    \begin{minipage}{0.6\textwidth}
        \centering
        % Add any text or content here if required
    \end{minipage}
\end{figure}

Online platforms have been increasingly prominent in disseminating harmful content, challenging the creation of safe digital spaces 
% \cite{alatawi2021survey}. 
\cite{alatawi2021survey,kursuncu2019predictive}. 
According to the PEW Research Center, 41\% of Americans report experiencing online harassment \cite{vogels2021state}, and recent research highlights a surge in religious hate speech during the COVID-19 pandemic \cite{10.1145/3465336.3475111}. Detecting toxic content, especially nuanced and context-dependent multimodal forms, such as memes, remains a complex task 
\cite{Kiela2020, sheth2022defining}.
% \cite{Kiela2020}. 
\citet{pavlopoulos2020toxicity} noted that context availability alters the perceived toxicity in 5.2\% of posts. Memes that appear harmless can convey harmful messages through sarcasm, irony, or cultural references, thus potentially reinforcing negative stereotypes or societal norms \cite{waseem-hovy-2016-hateful, benmoussa2020internet, doi:10.1177/14614448231198169}. Hence, accurate toxicity assessment requires a thorough understanding of context, as implicit evaluations can misinterpret tone or intent 
\cite{kursuncu2021bad,sheth2022defining}.
% \cite{sheth2022defining}.

Although online multimodal toxicity detection has witnessed improvements, current methods, such as PromptHate \cite{cao-etal-2022-prompting} and HateCLIPper \cite{Kumar2022}, rely solely on training data and pre-trained models, which limits their ability to capture complex contextual cues and nuances of toxicity. Moreover, while larger models, such as Flamingo \cite{Alayrac2022} and LENS \cite{berrios2023towards}, achieve comparable performance, they require substantial computational resources to deploy in limited-resource settings. These models often struggle to effectively address nuanced or context-dependent toxicity due to their reliance on pattern recognition without incorporating broader contextual knowledge \cite{pavlopoulos-etal-2022-detection}.

Our approach, \emph{Knowledge-Infused Distilled Vision-Language Model (KID-VLM)}, addresses these limitations by leveraging ConceptNet \cite{Speer2017}, a common sense knowledge graph (KG) that enhances the comprehension of contextual nuances.
In this study, the following research questions guide our investigation: \\ 
\textbf{RQ1:} How can we incorporate Knowledge Distillation (KD) from LVLMs and common sense knowledge from KGs into compact models, enhancing their effectiveness in detecting toxicity within memes? \\ 
\textbf{RQ2:} How does the generalizability of KID-VLM compare to baseline methods in detecting toxicity across datasets?
 \emph{HatefulMemes} \cite{Kiela2020} and \emph{HarMeme} \cite{pramanick-etal-2021-momenta-multimodal}? \\
 \vspace{-0.05em}
We present a novel approach that enhances \emph{compact} VLMs by combining KD from LVLMs with knowledge infusion from KGs, enriching them with both implicit and explicit contextual knowledge (see Figure \ref{fig:kid_vlm_motivation}). Utilizing CLIP \cite{radford2021learning} as the backbone, we extract visual and textual features from memes, supplemented with captions generated by the LLaVA 1.6-NeXT model \cite{liu2024llavanext} for additional contextual features. These features are then fused using Align Fusion \cite{kumar2022hate} to create a multimodal representation, refined with knowledge distilled from the teacher model's caption representations through consistency loss. This process enables the student model to learn implicit contextual cues. Using graph-based reasoning, we further incorporate external knowledge from ConceptNet, constructing a joint graph that combines the meme context with KG entities. This knowledge-enhanced multimodal representation is then optimized using cross-entropy loss for final toxicity predictions (see Figure \ref{fig:architecture}).

% \textbf{Key Contributions}
Our findings demonstrate the effectiveness of the \emph{KID-VLM} framework, which outperforms the baselines. Error analysis and ablation studies further highlight the significance of explicit and implicit context incorporated through KD and infusion, providing improvements of 10.6\% and 0.5\% in F1 and AUC, respectively. Further, our approach demonstrates better generalizability compared to baselines, as seen in our performance on the HarMeme Dataset, where KID-VLM outperforms all baselines with an AUC of 92.98.

\section{Related Work} 
\label{sec:related}
\subsection{Online Toxicity Detection}
Early approaches to online toxicity detection primarily focused on textual content, employing traditional machine learning techniques \cite{leo_enhancing_2023,saha_hate_2023}. These techniques have evolved to utilize deep learning in addressing its nuanced, context-dependent nature \cite{jonathan_feature_2023,karim_multimodal_2022}. 
\citet{kursuncu2019modeling} incorporated multiple dimensions of online content, utilizing domain-specific corpora in detecting malicious actors.
Multimodal Models emerged {\cite{Kumar2022},\cite{cao-etal-2022-prompting}}, leveraging pre-trained VLMs and prompt tuning to improve performance.  However, these methods lack mechanisms to integrate explicit, structured reasoning, such as socio-cultural norms, necessary for identifying subtle forms of hate in memes. Large multimodal systems, including Flamingo-80B \cite{Alayrac2022} and LENS \cite{berrios2023towards}, demand high computational costs, challenging deployment \cite{erol2025playing}. Recent work, such as Pro-Cap \cite{cao2023pro}, has attempted to address this by freezing VLMs and aligning textual-visual features for efficient inference. However, it still solely relies on training data and retains dependence on monolithic models without external knowledge. Efforts to improve interpretability included \cite{lin-etal-2023-beneath}, which distilled reasoning from LLMs into smaller architectures and a multimodal debate framework \cite{lin2024towards} that enhances transparency in meme analysis. 

% \subsection{Knowledge-Infused Learning (K-IL) \& Knowledge Distillation (KD)}
% \subsection{Knowledge-Infused Learning (K-IL)}
\subsection{Knowledge Enhanced Learning}
Knowledge Enhanced Learning can be broadly categorized into two complementary approaches: (i) Knowledge Infusion (KI), which integrates structured external knowledge, such as KGs and ontologies, into models, and (ii) Knowledge Distillation (KD), where a large teacher model guides the training of a compact student model, helping to achieve similar performance with lower computational cost.

\paragraph{KI from KGs.}
KI enhances the model's representations by incorporating external structured knowledge from KGs and ontologies to capture explicit context across modalities 
\cite{zhang2022greaselm,kursuncu2019knowledge,khandelwal2024domain}. 
% \cite{zhang2022greaselm}. 
\citet{xu2024knowledge} have improved synthetic clinical data generation by combining KGs with LLMs, enhancing performance, and addressing privacy and fairness concerns. \citet{lymperaiou2024survey} demonstrated how KI can improve common sense and temporal reasoning in multimodal contexts. Further, \citet{agrawal2023can} investigated how KGs can reduce hallucinations in LLMs, while \citet{padhi2024enhancing} enhanced cross-modal contextual congruence for crowdfunding campaigns. 

\paragraph{KD from Large Models.}
% KD process involves transferring knowledge from a larger teacher model to a smaller, more efficient student model. \citet{wang2020multimodal} demonstrated that KD effectively handles incomplete modalities by distilling information from an equipped multimodal system  \citet{padhi2025calibrating} demonstrated how external feedback from visual grounding models can be used to calibrate multimodal LLMs, showing that incorporating such cross-modal signals from LLMs improves model confidence and robustness.
KD is a process in which knowledge is transferred from a larger teacher model to a smaller, more efficient student model. \citet{wang2020multimodal} demonstrated that KD can effectively address incomplete modalities by distilling information from a fully equipped multimodal system. Building on this idea, \citet{padhi2025calibrating} demonstrated that external knowledge feedback enhances grounding, thereby improving calibration of multimodal LLMs. This highlights how incorporating cross-modal signals enhances model confidence and robustness. Similarly, \citet{hong_research_2023} applied KD to Graph Neural Networks (GNNs) to enhance the capturing of community structures. \citet{setiawan2024accurate} introduced an $n$-best reranking method to refine sequence-level KD for neural machine translation.  In contrast, \citet{gholami2024gold} proposed an out-of-distribution-guided framework that enhances distilled models through targeted data generation. Further, \citet{wei2024sentence} compared sentence-level and token-level distillation under optimal conditions for each approach. 

In contrast to prior work, our approach unifies KD with KI via a neurosymbolic framework that infuses explicit relational semantics from common-sense KGs and distills implicit knowledge from LVLMs. This approach aims to equip compact models with the ability to learn from both knowledge bases and large pre-trained models, thereby boosting their multimodal reasoning capabilities.

\begin{figure*}[hbt!]
 % \vspace{-4em}
    \centering
  \includegraphics[width=16.25cm]{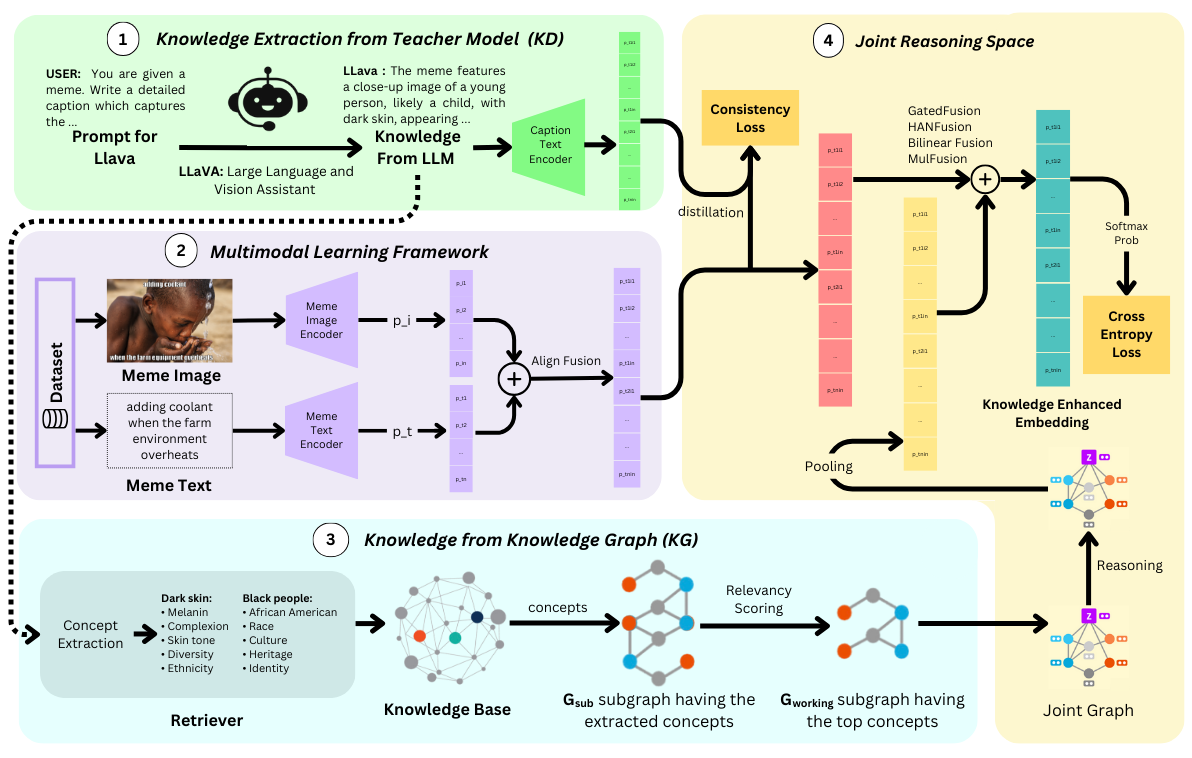}
\vspace{-0.8em}
\caption{\textbf{KID-VLM} framework: The framework unifies \ KD from an LVLM with KI from external KGs such as ConceptNet. The input image and text are processed through the CLIP encoders to generate embeddings, which are fused using different fusion mechanisms.  \textcircled{\tiny{1}} Knowledge Extraction from Teacher Model. \textcircled{\tiny{2}} Multimodal Learning Framework. \textcircled{\tiny{3}} Knowledge Extraction from KG. \textcircled{\tiny{4}} Joint Reasoning Space: reasoning using the implicit knowledge from teacher model and explicit knowledge from KG for toxicity prediction.
}
\vspace{-1em}
\label{fig:architecture}  
\end{figure*}
% \section{Preliminaries}
% \label{sec:prelim}
% \input{3-Preliminaries}
% \section{Exploratory Analysis}
% \label{sec:exploratory}
% \input{4-ExploratoryAnalysis}

% \vspace{-2mm}
\section{Methodology}
\label{sec:methodology}
% \vspace{-2mm}
Our approach, KID-VLM, illustrated in Figure \ref{fig:architecture}, utilizes a frozen Vision-Language (VL) Encoder as the student model \(S\), ConceptNet as the KG \(K\), and an LVLM as teacher model \(T\), serving as non-parametric knowledge sources to query explicit and implicit knowledge, respectively. For each data point, a meme with an image \(I_i\) and overlaid text \(T_i\), a joint multimodal feature representation \(s_i\) is generated using the pre-trained student model, \(s_i = S(I_i, T_i)\). To get the joint representation, we use the Align Fusion for it's low computational resource conditions \cite{kumar2022hate}.
% To get the joint representation, we use the Align Fusion Mechanism for its lower computational resource requirement compared to Cross Fusion \cite{kumar2022hate}.  
Concurrently, the teacher model \(T\) produces a caption \(C_i\) capturing the meme's implicit context, which is used for two purposes: (i) aid in teaching the student model through KD loss \( \mathcal{L}_{KD} \), and (ii) query ConceptNet to extract relevant sub-KGs $G_{sub}$ for each data point. The distilled representation $h_{distilled}$ from the student model is then fused with the pooled representation $h_{graph}$ from sub-KGs obtained through a GNN $G$, \(h_{graph} = G(G_{sub})\). This fused representation is optimized through Cross Entropy loss $L_{CE}$, enabling the model to synthesize outputs from both the teacher's latent reasoning and the explicit relational knowledge from the KG.

% The subsequent Section \ref{methodology-subsec:mutlimodal-rep-distilled} describes the creation of the distilled multimodal features $h_{distilled}$, and Section \ref{methodology-subsec:graph-based-reasoning} explains how external knowledge in the form of sub-KG \(G_{sub}\) is retrieved and integrated with $h_{distilled}$ to predict toxicity. 

% \vspace{-2mm}
\subsection{Knowledge Distillation} % for Multimodal Representation}
\label{methodology-subsec:mutlimodal-rep-distilled}
% \vspace{-1mm}
We utilize HateCLIPper \cite{kumar2022hate} as our frozen V\-L Encoder for stable feature extraction from each meme, while the teacher model, LLaVA-NeXT \cite{liu2024llavanext}, generates captions that capture the implicit context of the memes (See Appendix Table \ref{tab:llava_prompts}) KD is employed only during the training stage to optimize the student model's multimodal representation, \( \mathbf{s}_i \), by minimizing the consistency loss \( \mathcal{L}_{KD}\), defined as the Euclidean distance between the student's \( \mathbf{s}_i \) and the representations of the captions obtained from the teacher model. \( \mathbf{w}_i^{LLaVa} \):

\vspace{-1mm}
\begin{equation}
\mathcal{L}_{KD} = \| \mathbf{s}_i - \mathbf{w}_i^{LLaVa} \|_2^2
\end{equation}
\vspace{-1mm}

% where \(\mathbf{w}_i^{LLaVa}\) is the teacher model embedding.

This KD process, which does not extend to the testing stage, aligns the internal features of the student model with those of the teacher model, enabling the student to emulate the teacher's latent reasoning and recognize subtle implicit contextual cues in the memes. Note that LLaVA is not used at inference time. It is solely used in preprocessing to create captions that guide knowledge retrieval, ensuring our compact student model maintains a lightweight inference footprint and enabling deployment in low-resource settings.

% \vspace{-2mm}
\subsection{Graph-Based Reasoning}
\label{methodology-subsec:graph-based-reasoning}
% \vspace{-1mm}
To enrich the distilled multimodal representation with explicit relational knowledge, we employ a joint reasoning approach, which constructs joint \emph{working graphs} \(G_{\text{working}}\) for each data point, providing a reasoning space for the meme's overlaid text \(T_i\), sub-KG, and the generated caption \( C_i \). Then, this process generates a knowledge representation for working graphs through GNNs \cite{yasunaga2021qa}. 

% \vspace{-2mm}
\subsubsection{Working Graph Construction}
% \vspace{-1mm}
To construct the working graph \(G_{\text{working}}\), we first extract a sub-KG \( G_{\text{sub}} \) from ConceptNet (See Figure~\ref{fig:architecture}) incorporating concepts from the meme's text and the teacher-generated caption. We explicitly connect this sub-KG in a joint graph with a new context node \(z\), which represents the meme context and connects to each entity from the meme's text and teacher-generated text. The heuristic retrieval of \(G_{\text{sub}}\) from a KG can introduce irrelevant nodes, potentially adding noise and complicating reasoning, especially with large sub-KGs \citep{yasunaga2021qa}. 
% To address this, we apply two relevance scoring methods to assess the semantic alignment between the context node \(z\) and each candidate node \(v\) in \(G_{\text{sub}}\): one using the Roberta model \cite{liu2019roberta} with perplexity score and another using MiniLM \cite{wang-etal-2021-minilmv2} with cosine similarity. 

To assess the semantic alignment between the context node \(z\) and each candidate node \(v \in G_{\text{sub}}\), we apply two relevance scoring methods. The first approach uses the RoBERTa language model~\cite{liu2019roberta}, where the perplexity score of the concatenated textual inputs \([z; v]\) is used to evaluate relevance. The second utilizes MiniLM~\cite{wang-etal-2021-minilmv2} in conjunction with Sentence Transformers to compute the cosine similarity between the contextual and entity embeddings. For all main experiments and benchmark evaluations, we use MiniLM-based scoring due to its computational efficiency and strong performance. However, to evaluate robustness, we additionally report results using RoBERTa-based scoring in a separate ablation study (see Table~\ref{tab:roberta-results}).

The nodes within this sub-KG are ranked by relevance scoring to minimize noise and focus on the most relevant KG entities. This scoring involves evaluating the relationship of each node \( v \in V_{\text{sub}} \), where \(V_{\text{sub}} \) represents the nodes of \( G_{\text{sub}} \), with the overall meme context, encapsulated by the context node \( z\). The relevance score \( \rho_v \) is calculated through the Roberta model \cite{liu2019roberta} by passing the concatenated text representations of \( z \) and \( v \). The relevance score is determined using the perplexity score:

\vspace{-1em}
\begin{equation}
\rho_v = f_{\text{head}}(f_{\text{enc}}([\text{text}(z); \text{text}(v)]))
\end{equation}

where \( f_{\text{enc}} \) represents the encoder (Roberta) and \( f_{\text{head}} \) is the scoring head. A lower perplexity score indicates higher semantic relevance, helping prioritize nodes that align closely with the meme’s context. For additional validation, we compute the cosine similarity between the concatenated text representation of \( z \) and KG entities \( v \) using MiniLM \cite{wang-etal-2021-minilmv2} (Sentence-Transformer) as the relevance score \(\rho_v = cosine(z, v)\) to help identify the contextually significant nodes. 
We separately assess the relevance using scores from both Roberta and MiniLM in our experiments, selecting the top $k = 750$ most relevant entities to include in the working graph \( G_{\text{W}} \) (referred to as \( G_{\text{working}} \) in Figure~\ref{fig:architecture}). After constructing \( G_{W} \), context node \( z \) is added to bridge knowledge from the teacher model with that from KG. This node is then linked to the entities \( V_{\text{sub}} \)  in the working graph with a new relationship \( r' \), finalizing \( G_{W} \) for subsequent knowledge representation learning.

\definecolor{mydarkgreen}{rgb}{0.0, 0.5, 0.0}
\begin{table*}[h]
\centering
\small
\renewcommand{\arraystretch}{1.1}
\setlength{\tabcolsep}{2.5pt}
\resizebox{\textwidth}{!}{%
\begin{tabular}{@{}l|cc|cc|cc|cc|cc@{}}
\toprule
\textbf{Framework} & \multicolumn{2}{c}{\textbf{Accuracy}} & \multicolumn{2}{c}{\textbf{F1}} & \multicolumn{2}{c}{\textbf{Precision}} & \multicolumn{2}{c}{\textbf{Recall}} & \multicolumn{2}{c}{\textbf{AUC}} \\ 
\cmidrule(lr){2-3} \cmidrule(lr){4-5} \cmidrule(lr){6-7} \cmidrule(lr){8-9} \cmidrule(lr){10-11}
 & \textbf{Seen} & \textbf{Unseen} & \textbf{Seen} & \textbf{Unseen} & \textbf{Seen} & \textbf{Unseen} & \textbf{Seen} & \textbf{Unseen} & \textbf{Seen} & \textbf{Unseen} \\ 
\midrule
MMBT & 68.80 & 70.85 & 67.61 & 64.73  & 71.25 & 70.55 & 68.44 & 64.47 & 77.84 & 77.12 \\
CLIP\textsuperscript{*} & 59.50 & 70.85 & 53.88 & 65.67 & 76.40 & 69.75 & 25.10 & 65.21 & 72.14 & 76.25 \\ 
CLIP & 60.30 & 62.80 & 57.10 & 55.50 & 69.62 & 58.45 & 33.67 & 56.19 & 68.30 & 57.78 \\ 
\midrule
LLaVa\textsuperscript{+} & 54.30 & 63.55 & 29.58 & 19.60 & 60.38 & 53.85 & 19.59 & 28.74 & - & - \\
PromptHate & 76.10 & 75.75 & 75.76 & 72.94 & 77.30 & 74.66 & 76.10 & 72.20 & 84.21 & 80.43 \\
ExplainHM + LLM Debate & 69.40 & 68.05 & 65.23 & 55.72 & 73.59 & 58.01 & 58.57 & 53.50 & 74.61 & 71.82  \\
Pro-Cap + Targeted Captions & 75.50 & 74.35 & 75.30 & 73.46 & 76.02 & 73.21 & 75.35 & \textbf{74.41} & 83.65 & 81.34 \\
HateClipper & 76.07 & 76.65 & 73.00 & 64.21 & 78.00 & 75.50 & 56.00 & 55.87 & 85.14 & 83.70 \\
RGCL & 78.30 & \textbf{77.65} & 76.23 & 67.20 & 82.27 & 74.71 & 71.02 & 61.07 & \textbf{86.72} & 84.46 \\
\midrule
% KID-VLM + Hop 1 \& Roberta & 78.20 & 76.85 & 78.12 & 74.16 & 80.36 & 75.96 & 73.47 & 73.37 & 86.02 & 83.50 \\

KID-VLM + Hop 1 \& MiniLM  & \textbf{78.70} & 77.00 & \textbf{78.63} & 73.87 & 80.57 & 76.69 & 74.49 & 72.91 &  86.11 & \textbf{84.86} \\

% KID-VLM + Hop 2 \& Roberta  & 76.30 & 77.00 & 76.07 & 73.58 & 76.94 & \textbf{77.16} & \textbf{76.14} & 72.53 &  86.11 & 84.72 \\

KID-VLM + Hop 2 \& MiniLM  & 76.30 & 76.85 & 75.95 & \textbf{74.29} & \textbf{82.52} & 75.84 & 65.51 & 73.56 & 85.48 & 83.99 \\

\bottomrule
\end{tabular}%
}
\begin{minipage}{16cm}
\vspace{0.18cm}
\caption{Performance Comparison of various models on the \textbf{\emph{HatefulMemes Dataset}} (seen \& unseen splits). Top values for each metric are expressed in (\textbf{bold}). \textsuperscript{*} represents Frozen CLIP encoders, \textsuperscript{+} represents zeroshot evaluation.}
\label{tab:model-performance-hatememe}
\end{minipage}
\vspace{-10pt}
\end{table*}
% \vspace{-6mm}

\subsubsection{Knowledge Representation Learning}

To process and learn the representation of the joint working graph \( G_{W} \), we employ the \emph{Relational Graph Convolutional Network (R-GCN)} \cite{schlichtkrull_modeling_2017}, which enhances node representations by applying relation-specific transformations to aggregate information from neighboring nodes through structured interactions:

\vspace{-1.8em}
\begin{equation}
  h_v^{(l+1)} = \sigma \left( \sum_{r \in R} \sum_{u \in \mathcal{N}_r(v)} W_r^{(l)} h_u^{(l)} \right)  
\end{equation}
\vspace{-1em}

\noindent where \( h_v^{(l+1)} \) is the updated representation of node \( v \) at layer \( l+1 \), \( \sigma \) is the activation function, \( W_r^{(l)} \) is the relation-specific transformation matrix, \( \mathcal{N}_r(v) \) represents the set of neighbors connected to \( v \) by relation \( r \), and \( h_u^{(l)} \) is the representation of a neighboring node \( u \) at layer \( l \). 
% We also use \emph{Graph Attention Network (GAT)} \cite{velickovic_graph_2018} for the best performing variant of KID-VLM  
In addition, we learn a pooled representation of the graph to generate a single vector representing the entire working graph using Mean Pooling (Appendix \ref{appendix-graph-pooling}).

\paragraph{Fusion Mechanism:}
The pooled graph representation \( h_{\text{graph}} \) is then fused with the distilled multimodal representation \( h_{\text{distilled}} \) through \emph{Gated Fusion} \cite{xiong2024gated} mechanism:

\vspace{-2mm}
\begin{align}
\mathbf{F}_{\text{multimodal}} = \mathbf{G} \odot h_{\text{graph}} + (1 - \mathbf{G}) \odot h_{\text{distilled}}
\end{align}
\vspace{-2mm}

\noindent where \( \mathbf{F}_{\text{multimodal}} \) is fused multimodal representation and \( \mathbf{G} \) is the gating mechanism given by:

\vspace{-3mm}
\begin{align}
\mathbf{G} = \sigma\left( \mathbf{W}_g \left( h_{\text{graph}} \parallel h_{\text{distilled}} \right) \right)
\end{align}
\vspace{-6mm}

\( \mathbf{W}_g \) is a learnable weight matrix, and \( \sigma \) is the sigmoid activation function. 

\paragraph{Loss Function:}
The model is trained using a joint loss function: 

\vspace{-2mm}
\begin{equation}
\mathcal{L}_{\text{total}} = \lambda_1 \mathcal{L}_{\text{BCE}} + \lambda_2 \mathcal{L}_{\text{KD}}
\end{equation}
\vspace{-2mm}

\noindent where \( \mathcal{L}_{\text{BCE}} \) is the Binary Cross-Entropy Loss for meme classification, and \( \mathcal{L}_{\text{KD}} \) is the Consistency Loss for distillation, ensuring alignment between the multimodal fusion and captions from the teacher model. The losses are weighted using hyperparameters, and the overall loss is minimized using the AdamW optimizer.

% \vspace{-2mm}
\subsection{Experimentation and Evaluation} 
% \vspace{-0.5em}
We conducted experiments on two benchmark datasets: HatefulMemes \cite{Kiela2020} and HarMeme \cite{pramanick-etal-2021-detecting}. Detailed statistics are provided in Appendix \ref{sec:datasets}. For evaluation consistency with prior research \cite{Kumar2022, cao-etal-2022-prompting}, we used AUC, F1 score, Precision, Recall, and Accuracy. The HatefulMemes dataset provided two distinct `Seen' and `Unseen' data splits to assess the generalization ability and robustness of each model. The datasets include text annotations for memes, utilized as the text modality, avoiding any influence from LVLM’s OCR limitations. \\
% \vspace{-1em}\\
\textbf{KI: } We employed ConceptNet for explicit contextual information. We leveraged multi-hop traversal to expand the model's contextual understanding, specifically experimenting with \emph{Hop 1} and \emph{Hop 2} expansions. We also varied the $top_k$ most relevant nodes, where \( k \) ranged from $250$ to $750$. This range was chosen because higher values exceeded resource limitations, while fewer than $250$ nodes compromised the depth of contextual information. \\
% \vspace{-1em}
\textbf{Fusion: } For integrating multimodal and graph representations, we explored several fusion mechanisms, including \textit{Gated Fusion}, \textit{Multiplicative Fusion}, \textit{Bilinear Fusion}, and \textit{Hierarchical Attention Fusion} (Appendix \ref{sec:fusion-mechanism}). To process the $G_w$, we utilized \textit{RGCN} for all our experiments based on our ablation study results.  

% \vspace{-2mm}
\subsection{Baselines}
% \vspace{-1mm}
To evaluate the performance of our KID-VLM approach, we benchmarked against several \emph{compact} VLMs from prior research, which include HateClipper \cite{kumar2022hate}, RGCL \cite{RGCL}, Pro-Cap \cite{cao2023pro}, Explain-HM \cite{lin2024towards}, Mr. Harm \cite{lin-etal-2023-beneath}, and PromptHate \cite{cao-etal-2022-prompting}.
% , from a well-established benchmark based on the HatefulMemes challenge\footnote{\url{https://paperswithcode.com/sota/meme-classification-on-hateful-memes}}. 
We also incorporated the Multimodal Bitransformers (MMBT) \cite{kiela_supervised_2020} and various implementations of CLIP-based models \cite{radford2021learning} to facilitate a broad comparative analysis. In addition, we fine-tuned only the projection layers of CLIP, designated as CLIP\textsuperscript{*}. The LLaVA-NeXT \cite{liu2024llavanext} was used in a zero-shot setting to compare with a Large VLM. Our primary baseline reference for all reported values is RGCL. 

% \vspace{-0.8em}
\subsection{Implementation Details}
% \vspace{-0.5em}

We utilized Optuna to optimize the model's hyperparameters. Our parameter sweep included GNN hidden dimensions, number of layers, learning rate, fusion methods, and dropout rates. GNN hidden and output dimensions ranged from $2$ to $512$ and $2$ to $1024$, respectively. Mapping layer dimensions were set between $2$ and $2048$. Learning rates ranged from $1e-10$ to $1e-2$, and dropout probabilities from $0.0$ to $0.9$. We also tuned the number of mapping and pre-output layers ($1$ to $5$), weight decay ($1e-8$ to $1e-1$), and the loss alpha ($0$ to $1$). The model was trained for $30$ epochs via the AdamW optimizer, with learning rate warm-up and linear decay. We used a batch size of $4$ on RTX 5000 and L40 GPUs with $32$ and $40$ GB of VRAM, respectively. Model performance was monitored on the validation set, and the best model was selected based on validation AUC. The total trainable parameter count in KID-VLM remains well under 1B, including the CLIP encoders, the graph reasoning modules (R-GCN), and fusion layers, aligning with the compact model threshold from prior work \cite{izsak2019training, houlsby2019parameter}, which positions KID-VLM as performant and efficient.

% \vspace{-0.6em}
\section{Results \& Discussion}
\label{sec:results-disscuss}
% \vspace{-0.8em}

\definecolor{mydarkgreen}{rgb}{0.0, 0.5, 0.0}
\begin{table*}[hbt]
\centering
\small
\resizebox{0.8\textwidth}{!}{%
\begin{tabular}{@{}lccccc@{}}
\toprule
\textbf{Framework} & \textbf{Accuracy} & \textbf{F1} & \textbf{Precision} & \textbf{Recall} & \textbf{AUC} \\ 
\midrule
% CLIP-Text\textsuperscript{*} & 64.97 & 39.38 & 32.49 & 50.00 & 78.77 \\ 
% CLIP-Image\textsuperscript{*} & 78.81 & 74.40 & 79.06 & 72.92 & 91.35 \\ 
CLIP\textsuperscript{*} & 70.62 & 58.01 & 73.79 & 59.55 & 90.80 \\ 
% CLIP-Text & 77.40 & 76.37 & 75.99 & 78.15 & 84.07 \\ 
% CLIP-Image & 85.59 & 84.63 & 87.80 & 85.06 & 92.46 \\ 
CLIP & 66.67 & 55.04 & 62.21 & 56.51 & 80.82\\ 
\midrule
LLaVa\textsuperscript{+} & 80.23 & 67.59 & 79.35 & 58.87 & - \\
PromptHate & 78.53 & 78.28 & 80.02 & 82.73 & 90.72 \\
Pro-Cap+Target-based Captions & 83.90 & 83.18 & 82.62 & 85.38 & 90.97 \\
ExplainHM+LLM Debate & 84.75 & 77.50 & 80.17 & 75.00 & 92.05 \\
Mr.Harm+Label Rationale & 74.29 & 62.76 & \textbf{85.83} & 63.31 & - \\
HateClipper & 83.33 & 79.29 & 70.19 & \textbf{91.13} & 91.90  \\
RGCL & 84.75 & 79.39 & 75.36 & 83.87 & 90.10  \\
\midrule
KID-VLM+Hop 1 \& MiniLM & 84.46 & 83.85 & 83.38 & 86.37 & 92.77  \\
KID-VLM+Hop 2 \& MiniLM & \textbf{85.03} & \textbf{84.40} & 83.86 & 86.81  & \textbf{92.98}  \\
\bottomrule
\end{tabular}%
}
\begin{minipage}{16cm}
\vspace{0.18cm}
\caption{Performance Comparison of various models on HarMeme Dataset across multiple metrics.  The top values for each metric are expressed in (\textbf{bold}). \textsuperscript{*} represents Frozen CLIP encoders, \textsuperscript{+} represents zeroshot evaluation.}
\vspace{-2.5em}
\label{tab:model-performance-harmemedual}
\end{minipage}
\end{table*}

Our analysis shows that \emph{KID-VLM}, which integrates external knowledge from ConceptNet and distills information from the LLaVA model, outperforms the competitive baseline models from prior research \cite{RGCL, cao2023pro, lin2024towards, cao-etal-2022-prompting} across both the \emph{HatefulMemes} and \emph{HarMeme} datasets (see Tables \ref{tab:model-performance-hatememe} and \ref{tab:model-performance-harmemedual}). On the \emph{HatefulMemes} dataset across variants, our approach outperforms by 3.2\% in F1 score for the seen split and 10.6\% for the unseen split. It also improves AUC by 0.5\% on the unseen split. For the \emph{HarMeme} dataset, it shows improvements of 6.3\% and 3.2\% in F1 and AUC, respectively. The KID-VLM with MiniLM and Hop 2 traversal demonstrate overall higher performance on both datasets. KID-VLM with Hop 1 models also perform competitively, often matching or outperforming other models. We observe that baseline models often exhibit a trade-off between precision and recall, whereas KID-VLM models show a notable balance, reflected in higher F1 and AUC scores. This robust performance across metrics suggests the ability to capture the underlying patterns of toxicity. Further, our models' performance on the unseen split of the HatefulMemes dataset highlights better generalization capabilities, suggesting that the broader contextual understanding afforded by Hop 2 allows the model to generalize beyond the training data.

% Harmeme: 1.01 AUC, 1.5 F1
% Hatefulmemes: 0.5 AUC (unseen), 1.1 F1 (unseen), 3.2 F1 (seen)
% KID-VLM Hop 2 with MiniLM seems to be best KID model.
% test seen f1, auc kidvlm - 78.63, 86.11
% test seen f1 rgcl - 76.23, 86.72 
% test unseen f1, auc kidvlm - 73.87, 86.11
% test unseen f1 rgcl - 76.23, 86.72 

% \vspace{0.3cm}
\begin{table*}[h]
\centering
\small
\renewcommand{\arraystretch}{1.1}
\setlength{\tabcolsep}{2.5pt}
\resizebox{\textwidth}{!}{%
\begin{tabular}{@{}l|cc|cc|cc|cc|cc@{}}
\toprule
\textbf{KID-VLM Variant (RoBERTa)} & \multicolumn{2}{c}{\textbf{Accuracy}} & \multicolumn{2}{c}{\textbf{F1}} & \multicolumn{2}{c}{\textbf{Precision}} & \multicolumn{2}{c}{\textbf{Recall}} & \multicolumn{2}{c}{\textbf{AUC}} \\ 
\cmidrule(lr){2-3} \cmidrule(lr){4-5} \cmidrule(lr){6-7} \cmidrule(lr){8-9} \cmidrule(lr){10-11}
 & \textbf{Seen} & \textbf{Unseen} & \textbf{Seen} & \textbf{Unseen} & \textbf{Seen} & \textbf{Unseen} & \textbf{Seen} & \textbf{Unseen} & \textbf{Seen} & \textbf{Unseen} \\ 
\midrule
KID-VLM + Hop 1 \& RoBERTa & 78.20 & 76.85 & 78.12 & 74.16 & 80.36 & 75.96 & 73.47 & 73.37 & 86.02 & 83.50 \\
KID-VLM + Hop 2 \& RoBERTa & 76.30 & 77.00 & 76.07 & 73.58 & 76.94 & \textbf{77.16} & \textbf{76.14} & 72.53 & 86.11 & 84.72 \\
\bottomrule
\end{tabular}%
}
\begin{minipage}{16cm}
\vspace{0.18cm}
\caption{Performance of \textbf{KID-VLM variants using RoBERTa-based relevance scoring} on the \textbf{\emph{HatefulMemes Dataset}} (seen \& unseen splits). These results complement the MiniLM-based variants in Table~\ref{tab:model-performance-hatememe}.}
\label{tab:roberta-results}
\end{minipage}
\vspace{-10pt}
\end{table*}

% \vspace{-0.6em}
\subsection{Performance on HatefulMemes Dataset}
% \vspace{-0.5em}

Our framework, utilizing both Hop 1 and Hop 2 traversals, demonstrates consistent robust performance on both seen and unseen splits of the \emph{HatefulMemes} dataset, outperforming baseline models. Specifically, the KID-VLM with Hop 1 and Gated Fusion achieved an F1 score of $78.63$, an AUC of $86.11$, and an accuracy of $78.70$. Notably, the Hop 2 variant achieved the highest recall at $76.14$ while maintaining robust accuracy and AUC, highlighting its capability to detect contextual nuances of toxicity. The KID-VLM models showcase a well-balanced trade-off between precision and recall. The integration of \emph{MiniLM} for relevancy scoring in the creation of a working graph improved performance, which can be attributed to MiniLM’s ability to refine entity selection, leading to a more semantically coherent working graph, enhancing graph-based reasoning by reducing sparsity, and improving relational inference. While MMBT, CLIP, and LLaVA show varied results, CLIP-based models underperform on the unseen split, particularly in precision and recall. HateClipper and RGCL provide competitive baselines, with RGCL displaying an AUC of $86.72$ on the seen split.

% \vspace{-0.5em}
\subsection{Performance on HarMeme Dataset}
% \vspace{-0.5em}

On the \emph{HarMeme} dataset, \emph{KID-VLM} with Hop 2 achieved the highest overall performance with an F1 score of \emph{84.40} and an AUC of \emph{92.98}, which represents the state-of-the-art for this dataset. The Hop 1 model also outperforms the baselines, demonstrating the efficacy of the KID-VLM framework integrating external knowledge for meme classification. The improvement between Hop 2 and Hop 1 indicates the capturing of broader contextual cues enhancing performance. Mr.Harm and HateClipper outperform in precision and recall, respectively, while both fall short in recall and precision, showcasing a trade-off. Pro-Cap and ExplainHM exhibit more balanced results, albeit not matching the overall performance of KID-VLM. RGCL notably underperforms on this dataset, in contrast to its performance on the seen split of the HatefulMeme dataset (see Table \ref{tab:model-performance-hatememe}).

% \begin{table}[hbt!]
% \centering
% \small
% {
% \begin{tabular}{@{}c c c@{}}
% \toprule
% \textbf{Framework} & \textbf{\# Nodes}  & \textbf{AUC} \\ 
% \midrule
% KID-VLM & 250  & 84.91 \\
% KID-VLM & 500  & 85.76 \\
% KID-VLM & 750  & \textbf{86.11} \\
% \bottomrule
% \end{tabular}%
% }
% \caption{Ablation study on the impact of node count in Hop-1 ConceptNet subgraphs on HatefulMemes dataset. Models use an RGCN encoder and Gated Fusion with CLIP as visual/text encoder. Performance of KID-VLM improves with increasing node count}
% \vspace{-2em}
% \label{tab:ablation-study-nodes}
% \end{table}

% \vspace{-2mm}
\subsection{Impact of Node Count}
% \textbf{Impact of Node Count on Model Performance:} 
% \vspace{-0.5em}
The ablation study, summarized in Table \ref{tab:ablation-study-fusion}, illustrates the impact of increasing node count from KG on the model performance in AUC. With a node count of $250$, the model achieves an AUC of $84.91$, which improves to $85.76$ with $500$ nodes. When the node count is raised to $750$, the model yields the highest AUC of $86.11$. This trend indicates that incorporating more knowledge in the model from ConceptNet enhances its ability to discern and utilize contextual relationships within memes, thereby improving its reasoning capabilities and overall performance. On the other hand, it is crucial to ensure the relevance of the nodes to maintain performance quality and avoid introducing noise.

\begin{table}[hbt!]
\centering
\small
{
\begin{tabular}{c | c c || c | c c}
\toprule
  \# & \textbf{Nodes} & \textbf{AUC} & \# & \textbf{Fusion}  & \textbf{AUC} \\ 
\midrule
1  & 250  & 84.91 & 1 & Gated  &  \textbf{86.11} \\
2 & 500  & 85.76 & 2 & Multiplicative  &  80.64 \\
3  & 750  & \textbf{86.11} & 3 & Bilinear  & 85.26 \\
-  & - & - & 4 & Hierarchical Att.  & 85.39 \\
\bottomrule
\end{tabular}%
}
\vspace{-2mm}
% \caption{Ablation studies on the KID-VLM framework for the HatefulMemes Dataset. The left section details the impact of varying node counts (250, 500, 750) using Hop-1 with RGCN and Gated Fusion, demonstrating improved performance with increased node count. The right section evaluates different fusion mechanisms using Hop-2 with 750 nodes and CLIP, highlighting Gated Fusion as the most effective method based on AUC.}
\caption{Ablation studies on KID-VLM for the HatefulMemes Dataset analyze node count and fusion methods. The left section shows that increasing nodes (250, 500, 750) with Hop-1, RGCN, and Gated Fusion improves performance. The right section evaluates fusion mechanisms with Hop-2, 750 nodes, and CLIP, identifying Gated Fusion as the most effective based on AUC.}
\vspace{-1.5em}
\label{tab:ablation-study-fusion}
\end{table}

\subsection{Fusion \& GNN Architecture} 
% \vspace{-2mm}
Our ablation study (Table \ref{tab:ablation-study-fusion}) demonstrates the effectiveness of \emph{Gated Fusion}, balancing multimodal and graph-based features to prevent any modality from biasing the fusion process. Gated Fusion utilizes a dynamic gating mechanism that adaptively weighs the information from multimodal and graph-based representations using a sigmoid-activated gate. This allows for minimizing noise from external knowledge and preserving contextually relevant information, essential for toxicity detection. Gated fusion outperforms others, such as multiplicative fusion, which may diminish key signals in sparse feature space, and bilinear fusion, which may introduce complexity and increase the risk of overfitting. Hierarchical Attention may risk overlooking critical subtle cues necessary for identifying nuanced toxicity. Further, our experiments with GCN and GAT suggest that GCN, with an AUC of $86.11$, provides better performance through uniform feature aggregation. In contrast, GAT, with an AUC of $85.75$, may be more prone to noise due to its attention mechanism (Refer Appendix \ref{appendix-gat}).

\vspace{-1mm}
\begin{table}[hbt!]
\centering
\small
{
\begin{tabular}{@{}c c c c c c@{}}
\toprule
\textbf{Framework}  & \textbf{Acc.}  & \textbf{F1} & \textbf{Pre.} & \textbf{Rec.} & \textbf{AUC}\\ 
\midrule
No Knowledge & 76.07 &  73.00 & 78.00 & 56.00 & 85.14 \\
KI-VLM & 78.10 &  77.63 & 78.09 & \textbf{77.55} & 85.35 \\
KD-VLM & 73.70 &  73.12 & 75.33 & 73.44 & 85.53 \\
KID-VLM & \textbf{78.70} & \textbf{78.63} & \textbf{80.57} & 74.49 & \textbf{86.11} \\
\bottomrule
\end{tabular}%
}
\vspace{-2mm}
\caption{Ablation study on the impact of KI and KD on HatefulMemes Dataset using Hop-1 (750 nodes) and MiniLM for Relevancy Scoring.}
\vspace{-4mm}
\label{tab:ablation-study-parts}
\end{table}

% \vspace{-3mm}
\subsection{Impact of KI and KD}
% \vspace{-2mm}
The ablation study shown in Table \ref{tab:ablation-study-parts} demonstrates the complementary benefits of KI and KD in enhancing toxicity detection. KI-VLM leverages external commonsense knowledge from ConceptNet, improving Recall by $38.48\%$, F1 score by $6.34$, and AUC by $0.25\%$ over the baseline VLM with no knowledge, highlighting the value of explicit contextual reasoning. Similarly, KD-VLM enhances Recall by $31.14\%$ and AUC by $0.46\%$, suggesting that distilled knowledge from LVLMs improves implicit contextual understanding. The KID-VLM framework, which combines both KI and KD, achieves the best overall performance with improvements in AUC of $1.14\%$, F1 score of $7.71\%$, and accuracy of $3.46\%$, showcasing the effectiveness of fusing explicit and implicit knowledge for robust multimodal toxicity detection. In addition to MiniLM-based relevance scoring (Table~\ref{tab:model-performance-hatememe}), we included results using RoBERTa-based scoring in Table~\ref{tab:roberta-results}, showing comparable performance across splits, with the Hop 2 variant demonstrating higher recall ($76.14$) on seen data and AUC of $84.72$ on unseen data.

% \vspace{-3mm}
\begin{figure}[hbt!]
    \centering    \includegraphics[width=0.4\textwidth]{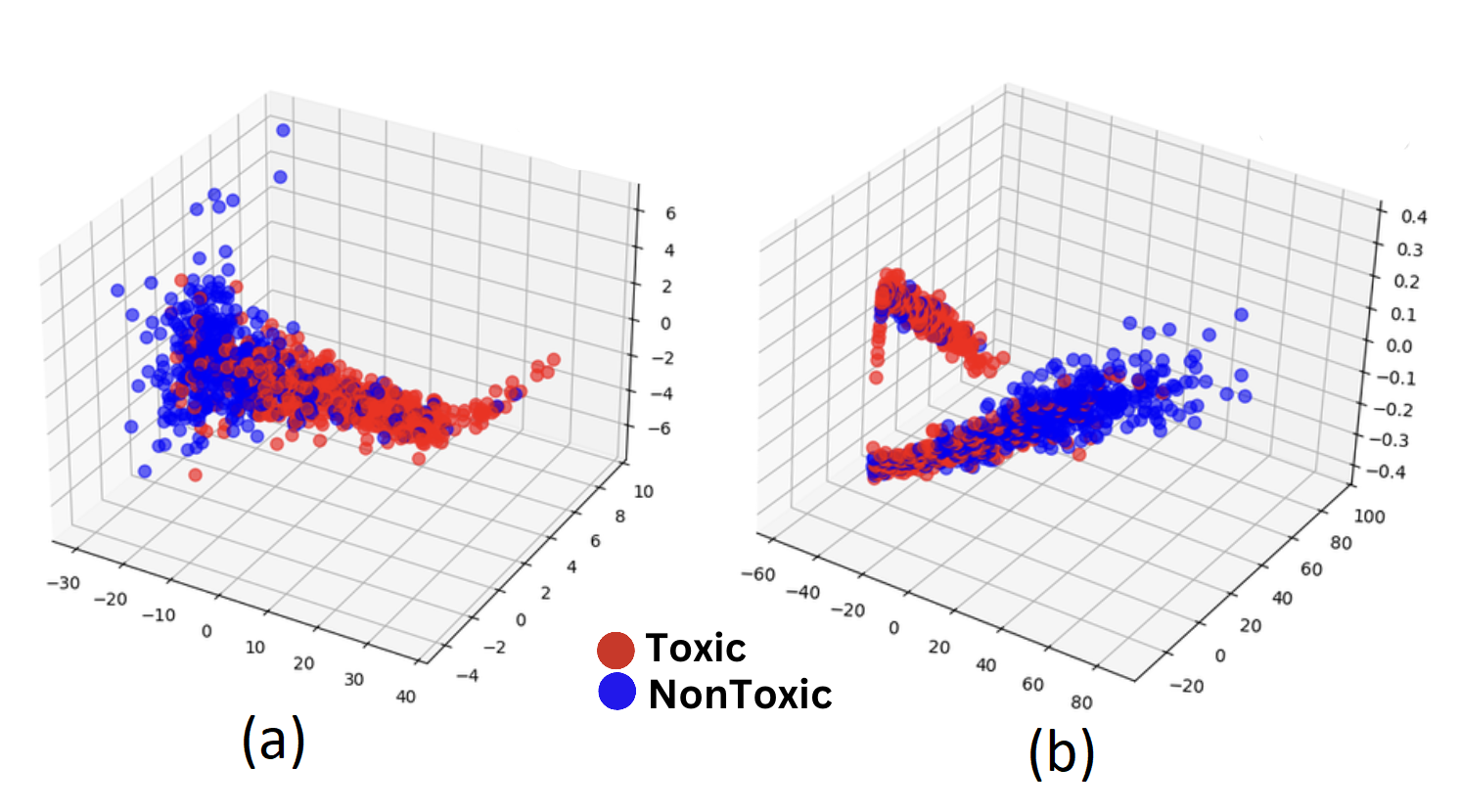} % Provide the path to your image
    \caption{Baseline (without KI/KD) vs. KID-VLM (with KI/KD) t-SNE plots illustrating the reduced 3D representation of the dataset after dimensionality reduction. The colors represent the ground truth labels of the data points. KID-VLM's plot shows a much clearer separation between the labeled data points.}
    \vspace{-4mm}
    \label{fig:pca_viz_baseline_kid}
\end{figure}

% \vspace{-0.8em}
\subsection{Discussion} 
% \vspace{-0.5em}
\paragraph{Better Separation: Toxic vs. Non-Toxic.} Enhancing representations by KI leads to a more distinct separation between toxic and non-toxic content as shown in Figure \ref{fig:pca_viz_baseline_kid}. Without knowledge-enhanced representations, the separation between these two classes tends to be ambiguous, and the data points are more scattered in the latent space. In contrast, with knowledge, there is a clearer separation through distinct regions with denser clusters.  This improved separation reduces the likelihood of misclassifications, which is often caused by ambiguity and unclear separation, especially in borderline cases where the context is crucial for detecting toxicity in a meme. This clearer separation is likely to stem from both explicit and implicit contextual cues learned during the training.

\vspace{-2mm}
\paragraph{Efficient Training and Deployment.} 
KID-VLM minimizes training overhead by using LVLMs only during inference, avoiding large-scale multimodal model training in prior works \cite{lin2024towards, liu2024improved}. By distilling crucial knowledge into a compact model with \textasciitilde 500M parameters,  rather than fine-tuning a large model, our approach substantially lowers computational costs. This facilitates efficient deployment in real-world settings where compute resources are limited, making KID-VLM scalable for toxicity detection applications.

\paragraph{Multi-Hop and Enhanced Generalization.} Leveraging multi-hop traversal helps our model to capture extended contextual cues, improving the overall performance on both the HarMeme and HatefulMemes datasets. It achieves particularly high scores in F1 score and AUC for unseen data, indicating a robust ability to handle new and diverse content. This enhanced generalization likely stems from the broader contextual understanding that Hop 2 provides, which is essential for navigating complex, nuanced content.

\section{Conclusion}
% \vspace{-3mm}
The results demonstrate that the \emph{KID-VLM} architecture, through its combination of KGs and KD, provides superior performance for detecting hate speech in memes compared to baseline methods. By integrating explicit relational knowledge with implicit contextual cues, KID-VLM better captures sarcasm, cultural references, and multimodal nuances that often obscure toxicity. The use of \emph{ConceptNet subgraphs} and distilled multimodal representations allows the model to understand the complex semantics of multimodal content better, resulting in improved accuracy, F1, and AUC scores. By distilling knowledge into a compact model (\textasciitilde 500M parameters), KID-VLM achieves these gains without the computational burden of large-scale models, making it scalable for real-world deployment.

% talk about novelty
% reduce numbers 
% explain words the results
% interpretaibilty 
% hop1 vs hop2 

% \begin{table}[t]
% \centering
% \small
% \captionsetup{font=footnotesize}
% % \resizebox{\textwidth}
% {}
% {%
% \begin{tabular}{@{}lp{2.5cm}llllll@{}}
% \toprule
% \textbf{Framework}  & \textbf{Encoder} &  \textbf{AUC} \\ 
% \midrule
% KID-VLM & RGCN & \textbf{86.11} \\
% KID-VLM & RGAT & 85.75 \\
% % MMBT & RGCN & - & - & - & - & - \\
% \bottomrule
% \end{tabular}%
% }
% \begin{minipage}{\linewidth}
% \vspace{0.18cm}
% \caption{
% \footnotesize Ablation Study on the HateMEME Dataset evaluating different Graph Encoders for the KID-VLM framework.}
% \label{tab:ablation-study-knowledge}
% \end{minipage}
% \end{table}

% \begin{table}[hbt!]
% \begin{table}

% % \captionsetup{skip=0pt}
% \centering
% \small
% {
% \begin{tabular}{@{}c c c c c c@{}}
% \toprule
% \textbf{Framework}  & \textbf{Acc.}  & \textbf{F1} & \textbf{Pre.} & \textbf{Rec.} & \textbf{AUC}\\ 
% \midrule
% HateClipper & 76.07 &  \textbf{73.00} & 78.00 & 56.00 & 85.14 \\
% KI-HateClipper & 78.10 &  \textbf{77.63} & 78.09 & 77.55 & 85.35 \\
% KD-HateClipper & 73.70 &  \textbf{73.12} & 75.33 & 73.44 & 85.53 \\
% KID-VLM & 78.70 &  \textbf{78.63} & 80.57 & 74.49 & 86.11 \\
% \bottomrule
% \end{tabular}%
% }
% \caption{Ablation study on the effect of Knowledge Infusion and Distillation for KID-VLM on HatefulMemes Dataset using hop-1 ConceptNet (750 nodes) and MiniLM for Relevancy Scoring.}
% \vspace{-1.5em}
% \label{tab:ablation-study-fusion}
% \end{table}

% \section{Conclusion}
% \label{sec:conclusion}
% \input{7-Conclusion}

% \newpage
\section{Limitations}
\label{sec:limitations}
While \emph{KID-VLM} demonstrates strong performance on the \emph{HatefulMemes} and \emph{HarMeme} datasets, it has several limitations. The model’s reliance on ConceptNet may limit its generalizability to other datasets beyond the two that have been extensively examined in this study. Additionally, incorporating graph-based methods may increase computational complexity, which can affect scalability for larger datasets. The quality of KD from larger models could degrade when using smaller student models, and there remains a potential risk of bias from pretrained models and KGs. Additionally, KID-VLM may inherit hallucination issues from its use of LLaVA. We have presented and discussed a few failure cases in Appendix \ref{sec:failure_cases}. Future work could address these concerns by exploring more diverse datasets, improving scalability, and investigating bias mitigation strategies.

\section{Ethical Considerations}
Toxicity detection systems can mislabel content or users, particularly when processing complex inputs such as memes that involve irony or satire. The cultural and social context may not be fully understood by the model, so continuous refinement is essential to mitigate biases. The study utilizes publicly available, anonymized datasets; however, the potential misuse of such systems in surveillance or censorship remains a concern. Responsible usage, with clear guidelines to protect free expression, is critical. Additionally, the potential for LVLMs to propagate biases from their training data must be addressed through ongoing evaluation and the use of diverse data sources.
% \newpage

% % Entries for the entire Anthology, followed by custom entries
% \bibliography{anthology,custom}
% \bibliographystyle{acl_natbib}

% \newpage

% \appendix

% \section{Appendix}
% \label{sec:appendix}
% \input{Appendix}
\bibliography{custom_clean}
\bibliographystyle{acl_natbib}

\clearpage
\appendix
\section*{Appendix}

% \section{Additional Results}

\section{Methodology}
\label{appendix-gat}

\subsection*{Graph Attention Networks (GAT)}
GAT employs an attention mechanism to dynamically weigh neighbor contributions, computing attention coefficients \( \alpha_{vu} \) between nodes, focusing on the most important neighbor interactions for node updates:

\vspace{-1em}
\begin{equation}
\alpha_{vu} = \text{\small softmax}\left(\small \text{LeakyReLU}\left( \mathbf{a}^\top [Wh_v \parallel Wh_u] \right) \right)
\end{equation}
\noindent where \( \alpha_{vu} \) is the attention coefficient between node \( v \) and neighbor \( u \), \( \mathbf{a} \) is a learnable attention vector, \( W \) is the weight matrix applied to node features, \( [Wh_v \parallel Wh_u] \) denotes the concatenation of transformed features of nodes \( v \) and \( u \), and \(\text{LeakyReLU}\) is the non-linear activation function.

\begin{table}[H]
\centering
% \caption{ Ablation Study for Fusion Model on HateMEME Dataset}
\small
{%
\begin{tabular}{@{}p{3cm}ccc@{}}
\toprule
\textbf{GNN Architecture}  & \textbf{AUC}  & \textbf{Acc.} & \textbf{F1} \\
\midrule
GAT  & 85.76  & 76.80 & 74.45  \\
R-GCN  & 86.11 & 76.30 & 76.14   \\

\bottomrule
\end{tabular}%
}
\begin{minipage}{\linewidth}
\vspace{0.2cm}
\caption{Ablation study on the GNN Architecture used to create for the Graph embedding. Results are reported for the seen split of the HatefulMemes Dataset on the best performing KID-VLM variant (RoberTa , Hop 2 , 750 nodes and Gated Fusion mechanism)}
\label{tab: gnn-architecture-ablation}
\end{minipage}
\end{table}

\label{appendix-graph-pooling}
\noindent \paragraph{Graph Pooling:} Once the node representations \( h_v \) for all nodes \( v \in G_{\text{W}} \) are updated through graph encoding (e.g., R-GCN, GAT), we learn a pooled representation of the graph to generate a single vector representing the entire working graph using Mean Pooling: 

\vspace{-0.6em}
\begin{equation}
    h_{\text{graph}} = \frac{1}{|V_{\text{W}}|} \sum_{v \in V_{\text{W}}} h_v
\end{equation}

\noindent where \( h_{\text{graph}} \) is the aggregated graph representation, \( V_{\text{W}} \) represents the nodes in the graph, \( |V_{\text{W}}| \) is the total number of nodes, and \( h_v \) is the feature representation of node \( v \). This pooled representation condenses the graph's information into a single vector.

\section{Datasets}
\label{sec:datasets}

\subsection*{Hateful Memes Dataset}

\begin{figure*}[hbt!]
 % \vspace{-4em}
    \centering
  \includegraphics[ width=\textwidth]{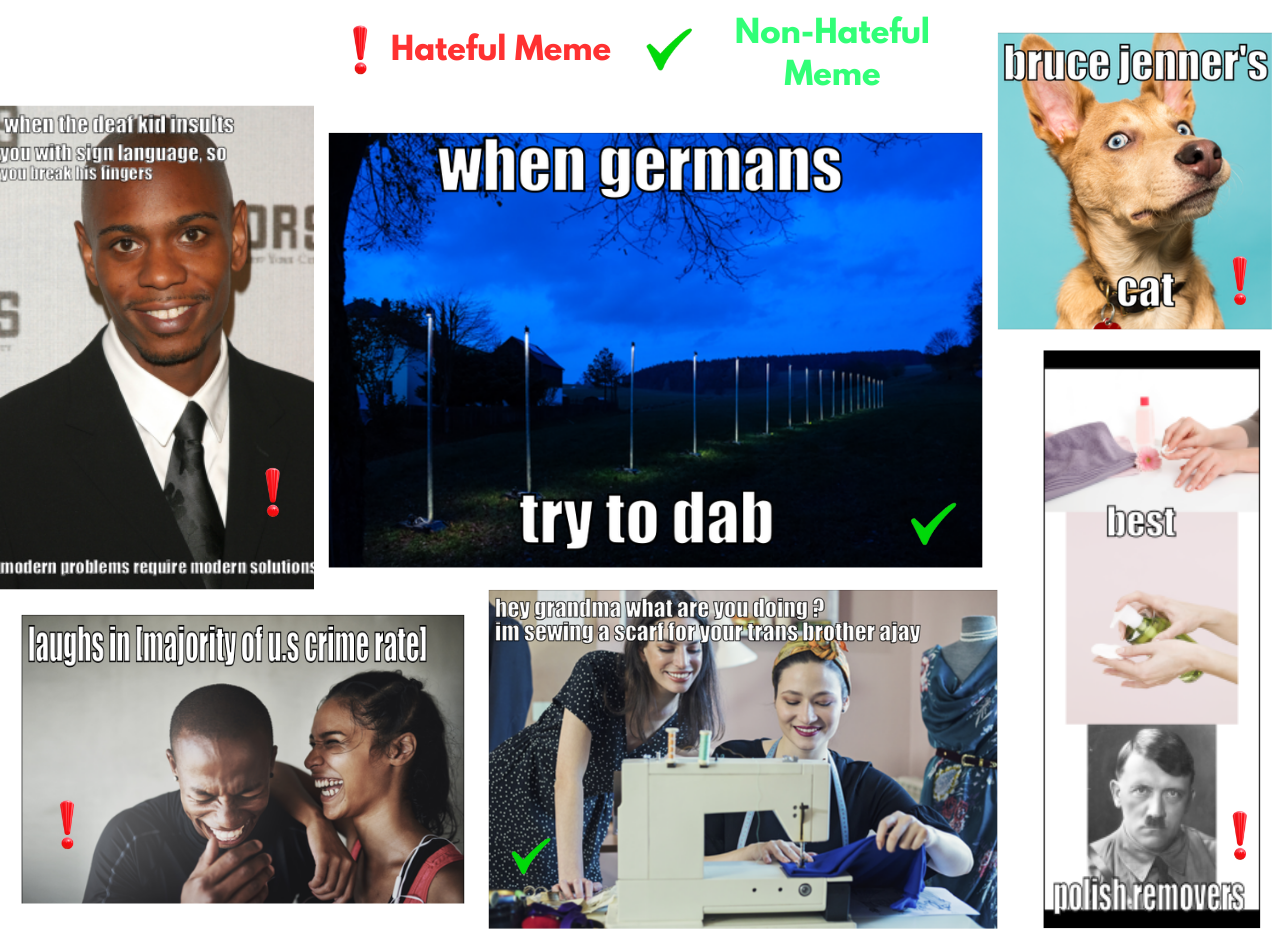}
\vspace{-0.8em}
\caption{ Examples from the Hateful Memes Dataset 
}
% \vspace{-1em}
\label{fig:examples-1}  
\end{figure*}

The primary dataset used in our experiments is the Hateful Memes Challenge Dataset \cite{Kiela2020}, consisting of 10K meme images accompanied by their respective text overlays. Each meme is labeled as either "Hateful" or "Non-hateful." The dataset is divided into training, testing, and validation subsets, as detailed in Table \ref{tab: hatememe-split}. Representative examples of memes from the dataset are presented in Figure \ref{fig:examples-1}.

\begin{table}[H]
\centering
% \caption{ Ablation Study for Fusion Model on HateMEME Dataset}
\small
{%
\begin{tabular}{@{}p{2cm}cccc@{}}
\toprule
\textbf{Labels}  & \textbf{Train}  & \textbf{Val} & \textbf{Test Seen} & \textbf{Test Unseen}\\ 
\midrule
Non-Hateful (0) & 5481 & 253 & 510  & 1250 \\
Hateful (1) & 3019  &247 & 490 & 750 \\
\textbf{Total} & 8500  & 500 & 1000 & 2000\\

\bottomrule
\end{tabular}%
}
\begin{minipage}{\linewidth}
\vspace{0.2cm}
\caption{Train, Test and Validation Seen splits for the Hateful Memes Dataset}
\label{tab: hatememe-split}
\end{minipage}
\end{table}

\subsection*{HarMeme Dataset}

\begin{figure*}[hbt!]
 % \vspace{-4em}
    \centering
  \includegraphics[ width=\textwidth]{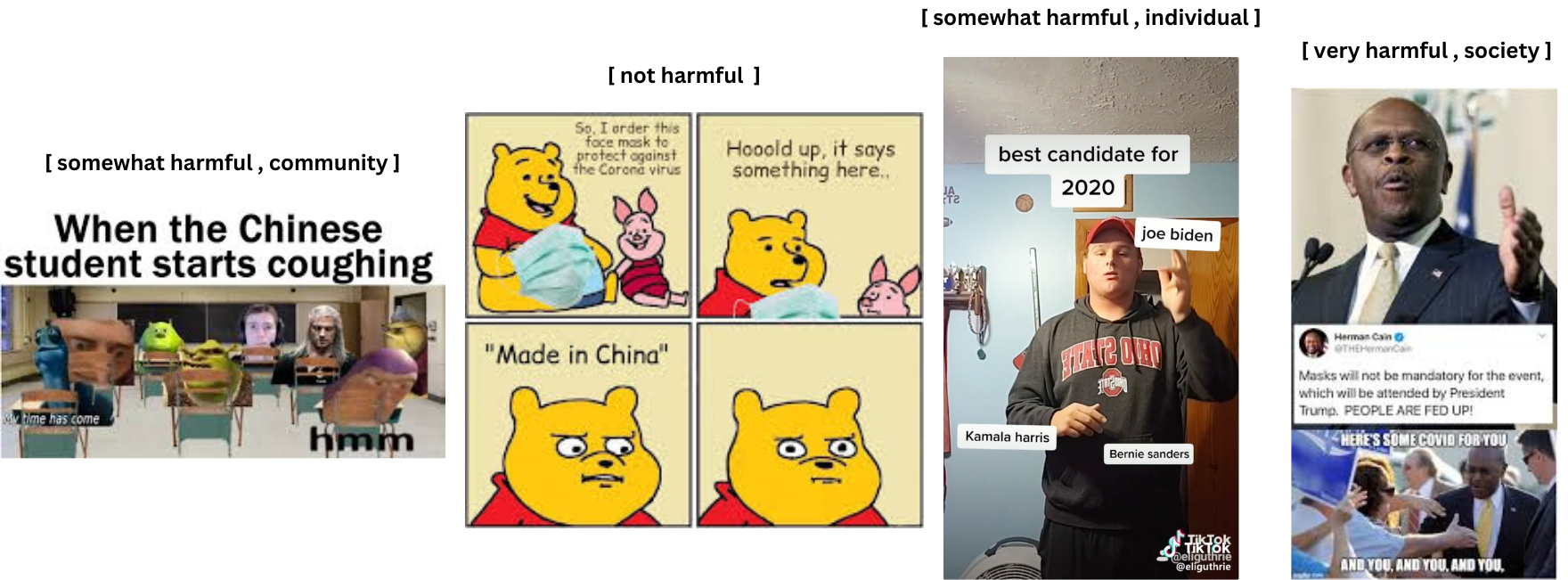}
\vspace{-0.8em}
\caption{ Examples from the HarMeme Dataset. The labels are given in the format [ Intensity , Target ] (Target label is not defined for not harmful memes)
}
% \vspace{-1em}
\label{fig:examples-2}  
\end{figure*}

We also conducted a series of experiments on the HarMeme Dataset \cite{pramanick-etal-2021-detecting}, a benchmark dataset for hateful meme classification. This dataset comprises 3,544 memes related to COVID-19, collected from the Internet, with each meme annotated with both [Intensity, Target] labels. The dataset is split into training, testing, and validation sets for the Intensity and Target labels, as detailed in Tables \ref{tab: harmeme-split-intensity} and \ref{tab: harmeme-split-target}, respectively. Figure \ref{fig:examples-2} presents several example memes from the dataset. As all the baselines studied use the 2-class version of the Intensity variable i.e. they combine the Somewhat Harmful and Very Harmful classes into a single Harmful class (Refer Table \ref{tab: harmeme-split-intensity-2}) as the Very Harmful class is under-represented in the dataset (\textasciitilde 6 \%)

\begin{table}[H]
\centering
% \caption{ Ablation Study for Fusion Model on HateMEME Dataset}
\small
{%
\begin{tabular}{@{}p{3cm}c c c@{}}
\toprule
\textbf{Labels}  & \textbf{Train Set}  & \textbf{Val Set} & \textbf{Test Set}\\ 
\midrule
Not Harmful (0) & 1949 & 116 & 230  \\
Harmful (1) & 1062  &61 & 124 \\
\textbf{Total} & 3013  & 177 & 354 \\

\bottomrule
\end{tabular}%
}
\begin{minipage}{\linewidth}
\vspace{0.2cm}
\caption{Train, Test and Validation splits for the HarMeme Dataset Intensity Variable for 2 label}
\label{tab: harmeme-split-intensity-2}
\end{minipage}
\end{table}

\begin{table}[H]
\centering
% \caption{ Ablation Study for Fusion Model on HateMEME Dataset}
\small
{%
\begin{tabular}{@{}p{3cm}c c c@{}}
\toprule
\textbf{Labels}  & \textbf{Train Set}  & \textbf{Val Set} & \textbf{Test Set}\\ 
\midrule
Not Harmful (0) & 1949 & 116 & 230  \\
Somewhat Harmful (1) & 882  &51 & 103 \\
Very Harmful (2) & 182  & 10 & 21 \\
\textbf{Total} & 3013  & 177 & 354 \\

\bottomrule
\end{tabular}%
}
\begin{minipage}{\linewidth}
\vspace{0.2cm}
\caption{Train, Test and Validation splits for the HarMeme Dataset Intensity Variable for 3 labels (Refer Table \ref{tab:model-performance-harmeme} )}
\label{tab: harmeme-split-intensity}
\end{minipage}
\end{table}

\begin{table*}[hbt]
\centering
\resizebox{16cm}{!}{
\begin{tabular}{@{}l|llll|llll@{}}
\toprule
\textbf{Framework} & \multicolumn{4}{c|}{\textbf{Target Metrics}} & \multicolumn{4}{c}{\textbf{Intensity Metrics}} \\ 
% \midrule
 &  \textbf{Acc.} & \textbf{F1}& \textbf{Precision} & \textbf{Recall} & \textbf{Acc.} & \textbf{F1}& \textbf{Precision} & \textbf{Recall} \\ 
\midrule
% MMBT & - & - & - & - & - & - & - & - & - \\
% \midrule
% CLIP-Text & 67.74 & 54.39 &41.33 & 37.61 & 75.14 & 50.09& 48.45 & 53.38 \\ 
% CLIP-Image &62.10& 36.41 & 60.57 & 39.24 & 80.23 & 53.54& 51.48 & 56.89\\ 
CLIP & 57.26 & 35.67 &50.29 & 39.24 & 80.23 & 53.60& 51.55 & 57.06 \\ 
% CLIP-Text\textsuperscript{*} & 59.68& 35.43 &74.42 & 52.17 & 68.36 & 45.13& 43.70 & 47.94 \\ 
% CLIP-Image\textsuperscript{*} & 62.10& 36.41 & 60.57 & 39.24 &64.97 & 26.26& 21.66 & 33.33  \\ 
CLIP\textsuperscript{*} & 62.10& 36.88 & 39.86 & 38.88 & 77.12 & 50.59 & 49.03 & 52.26 \\
PromptHate & 73.73& 72.74 & 75.32& 73.73 & 74.58 & 74.50 & 78.97 & 74.58 \\ 
HateClipper & 75.00 & 73.42& 75.83 & \textbf{79.66} & 79.66 & 78.76 & 75.83 & 79.66 \\
\midrule
KID-VLM (Hop 2) & \textbf{77.42}\scriptsize{\textcolor{mydarkgreen}{(+2.42)}} & \textbf{75.88}\scriptsize{\textcolor{mydarkgreen}{(+2.46)}}& \textbf{78.51}\scriptsize{\textcolor{mydarkgreen}{(+2.68)}} & 77.42\scriptsize{\textcolor{red}{(-2.22)}} & \textbf{81.07} \scriptsize{\textcolor{mydarkgreen}{(+1.41)}} & \textbf{79.08} \scriptsize{\textcolor{mydarkgreen}{(+0.32)}}& {77.73}\scriptsize{\textcolor{mydarkgreen}{(+1.90)}} & \textbf{81.07} \scriptsize{\textcolor{mydarkgreen}{(+1.41)}}\\
KID-VLM (Hop 1)& \textbf{77.42}\scriptsize{\textcolor{mydarkgreen}{(+2.42)}} & \textbf{75.88}\scriptsize{\textcolor{mydarkgreen}{(+2.46)}}& \textbf{78.51} \scriptsize{\textcolor{mydarkgreen}{(+2.68)}}& 77.42\scriptsize{\textcolor{red}{(-2.22)}} & 80.51 \scriptsize{\textcolor{mydarkgreen}{(+0.85)}} & 78.73 \scriptsize{\textcolor{red}{(-0.03)}}& \textbf{78.10} \scriptsize{\textcolor{mydarkgreen}{(+2.27)}}& 80.51 \scriptsize{\textcolor{mydarkgreen}{(+0.85)}} \\

\bottomrule
\end{tabular}
}
\begin{minipage}{16cm}
\vspace{0.18cm}
\caption{Performance Comparison of models on HarMeme Dataset across multiple metrics. The KID-VLM framework, using LLaVA for caption generation and MiniLM for relevancy scoring, outperforms other models (\textbf{bold}). \textsuperscript{*} represents Frozen CLIP encoders. The deltas are reported with respect to the HateClipper baseline.}
\vspace{-2em}
\label{tab:model-performance-harmeme}
\end{minipage}
\end{table*}
\begin{table}[H]
\centering
% \caption{ Ablation Study for Fusion Model on HateMEME Dataset}
\small
{%
\begin{tabular}{@{}p{3cm}c c c@{}}
\toprule
\textbf{Labels}  & \textbf{Train Set}  & \textbf{Val Set} & \textbf{Test Set}\\ 
\midrule
Individual & 493 & 30 & 59  \\
Organisation & 65  & 3 & 7 \\
Community & 279  & 16 & 32 \\
Society & 226  &13 & 26 \\
\textbf{Total} & 1063  & 62 & 124 \\

\bottomrule
\end{tabular}%
}
\begin{minipage}{\linewidth}
\vspace{0.2cm}
\caption{Train, Test and Validation splits for the HarMeme Dataset Target Variable (Refer Table \ref{tab:model-performance-harmeme} )}
\label{tab: harmeme-split-target}
\end{minipage}
\end{table}

% \newpage

\section{Fusion Mechanisms}
\label{sec:fusion-mechanism}
Let \( \mathbf{E}_g \in \mathbb{R}^{d_g} \) and \( \mathbf{E}_m \in \mathbb{R}^{d_m} \) denote the graph pooled representation and the distilled multimodal representation, respectively, where \( d_g \) and \( d_m \) represent the dimensions of the graph and multimodal embeddings. Several fusion mechanisms are evaluated for combining these embeddings from different modalities:

\paragraph{Gated Fusion}
Gated Fusion applies a gating mechanism to combine the graph and multimodal embeddings. The fusion is computed as:

\begin{equation}
\mathbf{F}_{\text{gated}} = 
\begin{aligned}
    &\sigma(\mathbf{W}_g [\mathbf{E}_g \| \mathbf{E}_m]) \odot \mathbf{E}_g \\
    &+ (1 - \sigma(\mathbf{W}_g [\mathbf{E}_g \| \mathbf{E}_m])) \odot \mathbf{E}_m
\end{aligned}
\end{equation}

where \( \mathbf{W}_g \in \mathbb{R}^{(d_g + d_m) \times d_g} \) represents learnable weights, \( \sigma(\cdot) \) is the sigmoid activation function, \( \odot \) denotes element-wise multiplication, and \( \| \) represents concatenation of the embeddings. 

\paragraph{Bilinear Pooling Fusion}
Bilinear Pooling Fusion computes the bilinear interaction between the graph and multimodal embeddings:

\begin{equation}
\mathbf{F}_{\text{bilinear}} = \mathbf{E}_g^T \mathbf{W}_b \mathbf{E}_m
\end{equation}

where \( \mathbf{W}_b \in \mathbb{R}^{d_g \times d_m} \) is a bilinear transformation matrix that models interactions between the modalities.

\paragraph{HAN Fusion}
Hierarchical Attention Network (HAN) Fusion applies attention weights at multiple levels to the graph and multimodal embeddings:

\begin{equation}
\mathbf{F}_{\text{HAN}} = \sum_{l=1}^{L} \alpha_l (\mathbf{W}_l [\mathbf{E}_g \| \mathbf{E}_m])
\end{equation}

where \( \alpha_l \) is the attention weight at level \( l \), \( \mathbf{W}_l \in \mathbb{R}^{(d_g + d_m) \times d} \) is the learnable weight matrix, and \( L \) is the total number of hierarchical levels.

\paragraph{Multiplicative Fusion}
Multiplicative Fusion combines the embeddings multiplicatively after non-linear transformation:

\begin{equation}
\mathbf{F}_{\text{mult}} = \tanh(\mathbf{W}_m \mathbf{E}_g) \odot \tanh(\mathbf{W}_m \mathbf{E}_m)
\end{equation}

where \( \tanh(\cdot) \) is the hyperbolic tangent activation function, and \( \mathbf{W}_m \in \mathbb{R}^{d_g \times d_m} \) is a weight matrix that transforms the embeddings before interaction.

\section{Interpretability Examples} 
\label{sec: interpret-appendix}
By incorporating ConceptNet,the model's interpretability is enhanced, enabling it to reason more effectively about the relationships between textual and visual elements in memes. The extracted nodes from ConceptNet offer semantic context that aids in understanding both literal and metaphorical meanings, improving the model's ability to explain its reasoning.\\

\begin{figure*}[hbt!]
 % \vspace{-4em}
    \centering
  \includegraphics[ width=\textwidth]{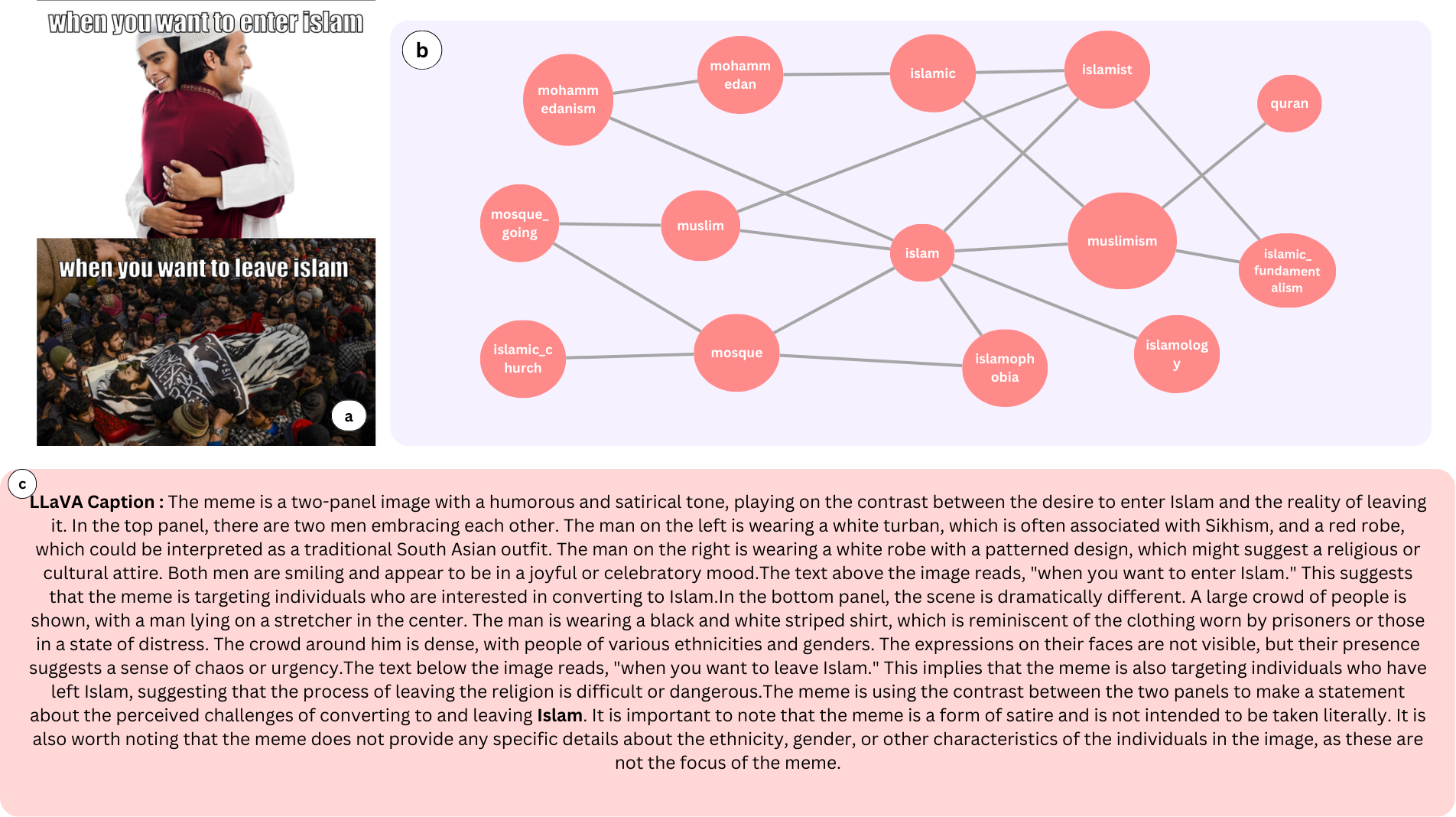}
\vspace{-0.8em}
\caption{ \textcircled{{a}} Example meme image from the Hateful Meme Dataset. \textcircled{b}  Portion of extracted subgraph after Relevancy Scoring and Pruning. \textcircled{c} Generated LLaVA caption for the meme 
}
\label{fig:interpretability-appendix-1}  
\end{figure*}

As illustrated in Figure \ref{fig:interpretability-appendix-1}
, the LLaVA caption interprets the meme as contrasting the perceived ease of entering Islam, depicted as a joyful experience, with the challenges of leaving Islam, which are portrayed as distressing or dangerous. The top panel shows a celebratory embrace, while the bottom panel illustrates chaos and urgency surrounding an individual in distress. The knowledge sub-graph, containing nodes such as Islamophobia, Muslim, and Islam, underscores the religious context. When combined, the LLaVA explanation and the knowledge graph reveal how the meme perpetuates harmful stereotypes and fosters negative sentiments towards Islam, reinforcing its toxicity.

\begin{figure*}[hbt!]
 % \vspace{-4em}
    \centering
  \includegraphics[width=16.2cm]{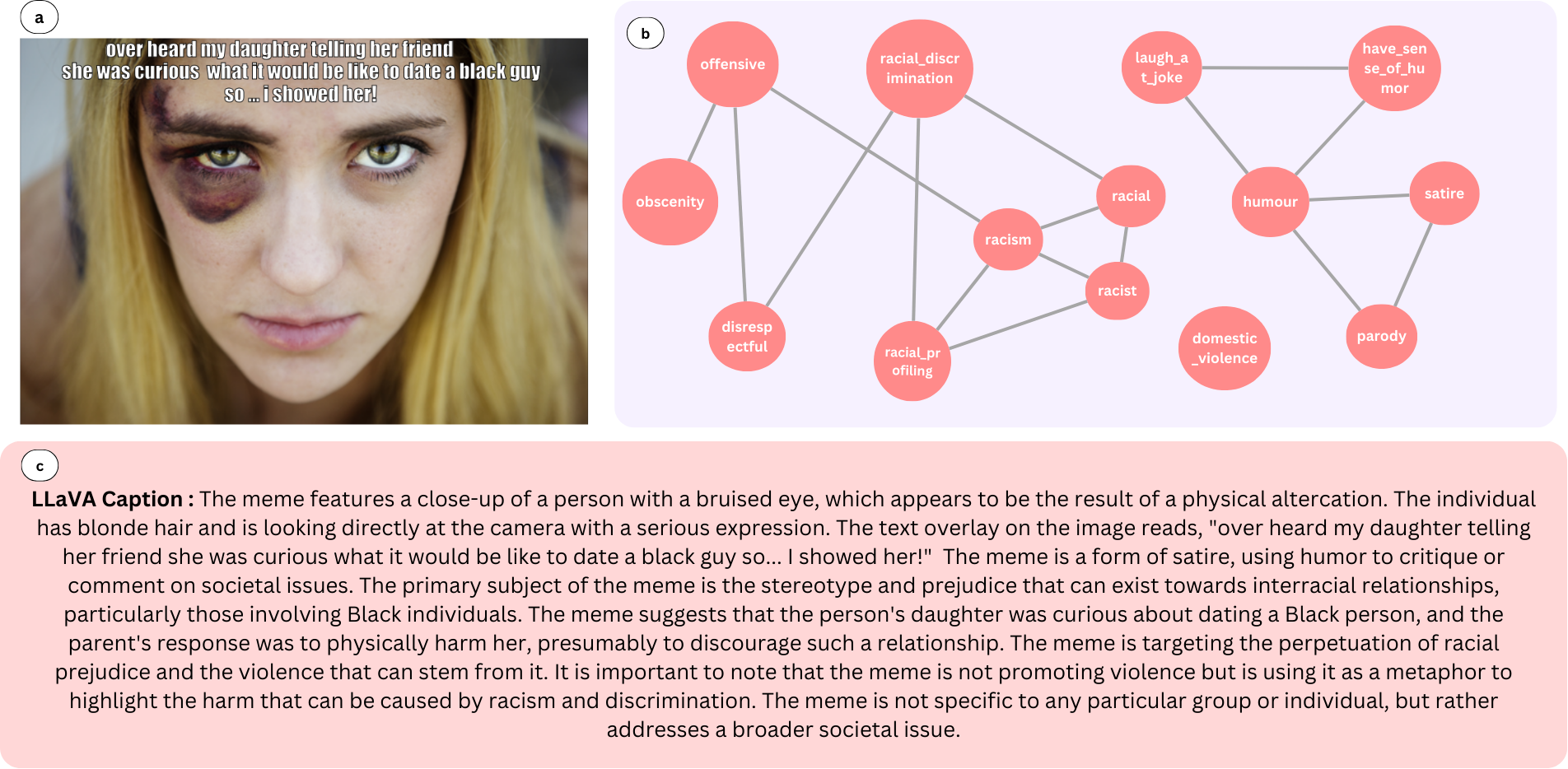}
\vspace{-0.8em}
\caption{ \textcircled{{a}} Example meme image from the Hateful Meme Dataset. \textcircled{b}  Portion of extracted subgraph after Relevancy Scoring and Pruning. \textcircled{c} Generated LLaVA caption for the meme 
}
\label{fig:interpretability-appendix-2}  
\end{figure*}

Similarly, in Figure \ref{fig:interpretability-appendix-2}, the LLaVA caption interprets the meme as addressing issues of racial discrimination and violence. The meme features a close-up image of a person with a bruised eye, implying a physical altercation. The individual, who has blonde hair, stares directly at the camera with a serious expression. The text on the image reads, "overheard my daughter telling her friend she was curious what it would be like to date a black guy so... I showed her!" The meme uses satire to comment on societal issues, specifically the harmful stereotypes and prejudices surrounding interracial relationships, particularly those involving Black individuals.The knowledge sub-graph includes nodes such as racial discrimination, racism, and domestic violence, underscoring the meme's connection to racial issues and violence. The meme critiques how prejudice and discrimination can manifest, especially when parents perpetuate such harmful views in response to their children's actions. While the meme employs dark humor and satire, it ultimately reflects on the damaging effects of racism and discrimination. It is important to note that the meme does not endorse violence but rather uses the image as a metaphor to highlight the harmful consequences of racial bias, addressing a broader societal issue.

The model's ability to visualize and comprehend the reasoning behind toxicity classification represents a significant advancement in developing more explainable AI systems for multimodal content analysis.

\vspace{0.4cm}
\begin{table*}[t]
\resizebox{\textwidth}{!}{
\begin{tabular}{p{4cm}p{12cm}}
    \toprule[1pt]
    \textbf{\centering Prompts for Dataset} & \textbf{\centering Prompt for Teacher LVLM} \\
    \toprule[1pt]
    Prompts for Captioning HateMeme & 
    \small \texttt{You are given a meme. Write a detailed caption which captures the context, emotions and the theme of the meme.  Identify and elaborate on the primary subject or target of the meme, which could include specific groups or individuals such as Muslims, Jews, transgender individuals, gays, Black people, Asians, etc.} \\
    \toprule
    Prompts for Captioning HarMEME & 
    \small \texttt{You are given a meme. Write a detailed caption that captures the context, emotions, and theme of the meme. Describe all visible elements in the image, including the ethnicity, gender, and other characteristics of individuals. Do not omit or censor any details.} \\
    \toprule
    
\end{tabular}
}
\caption{Prompts used to generate captions and analyze meme content using the LLaVA model.}
\label{tab:llava_prompts}
\end{table*}

\section{Failure Cases}
\label{sec:failure_cases}

Our model utilizes caption embeddings for knowledge distillation, which has led to a few misclassifications due to subtle nuances in image-text relationships. Below, we highlight three cases where the model's predictions differed from the ground truth.

\subsection{Case 1: Misclassification Due to Unexpected Context}
\label{subsec:case1}
\begin{figure}[h]
    \centering
    \includegraphics[width=0.6\linewidth]{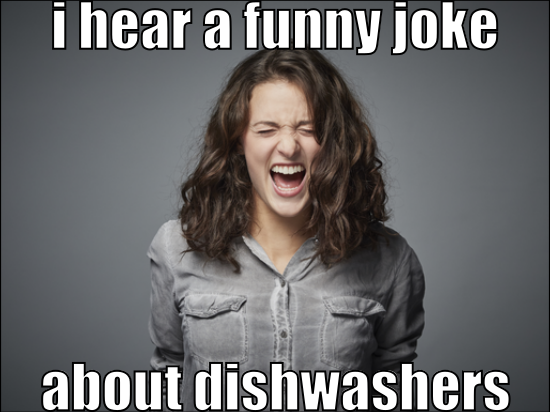}
    \caption{Meme with surprised woman. \\ \textbf{LLaVA caption:} The meme features a woman with curly hair, who appears to be in a state of surprise or shock. She is wearing a light-colored top with a collar, and her mouth is open as if she is in the middle of speaking or reacting to something. The background is a neutral, muted gray, which puts the focus on her expression. The text on the meme reads, \textit{"i hear a funny joke about dish washers."} The humor in the meme seems to be derived from the juxtaposition of the woman's surprised expression with the mundane and unexpected subject of the joke. The meme is likely intended to be humorous by playing on the idea that a joke about dish washers is not a typical topic for a joke, and the woman's reaction suggests that she is surprised by the subject matter. The meme does not target any specific group or individual, and there are no visible characteristics that would indicate the ethnicity or gender of the woman. The meme is focused on the reaction to the joke, rather than the identity of the person reacting. The primary subject or target of the meme is the unexpected and possibly absurd nature of the joke itself.}
    \label{fig:case1}
\end{figure}

The image in Figure~\ref{fig:case1} shows a woman with curly hair, wearing a light-colored collared top, and reacting with surprise. The caption reads: \textit{"i hear a funny joke about dish washers."} 

The model incorrectly classified this image as toxic. This misclassification likely occurred because jokes about dishwashers are uncommon, and in some cases, household appliances—especially dishwashers—have been used in derogatory jokes about women. Since the meme presents a reaction of surprise to such a joke, the model may have inferred that the joke carried a sexist connotation, even though there is no direct implication of that in the image or text. This suggests the model was influenced by prior associations rather than the actual content of the meme.

\subsection{Case 2: Failure to Recognize the Referenced Individual}
\label{subsec:case2}
\begin{figure}[h]
    \centering
    \includegraphics[width=0.6\linewidth]{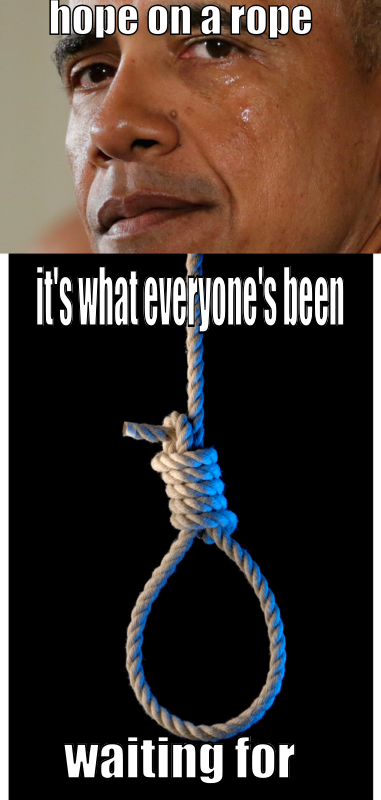}
    \caption{Meme with noose juxtaposition. \\ \textbf{LLaVA caption:} The meme features a split image with two distinct parts. On the left side, there is a photograph of a man with a serious expression. He appears to be middle-aged, with dark skin and short hair. The man is wearing a dark suit and has a slight frown on his face. On the right side of the meme, there is a graphic illustration of a noose with a blue rope. The noose is hanging against a black background, and the phrase \textit{"it's what everyone's been waiting for"} is written above it. The meme seems to be a form of satire or dark humor, using the juxtaposition of the man's serious expression with the noose to create a shocking or ironic effect. The phrase \textit{"it's what everyone's been waiting for"} suggests that the noose represents a long-awaited or anticipated event, which is likely meant to be a commentary on a specific political or social issue. The primary subject or target of the meme is not explicitly stated, but it appears to be directed towards a group of people who are perceived to be waiting for or advocating for a particular outcome. The meme's use of a noose as a symbol may be intended to provoke a strong emotional response, possibly to criticize or satirize the views of the targeted group. It is important to note that the meme contains potentially offensive content, and its use of a noose as a symbol may be insensitive or inappropriate. The meme's message and intent are open to interpretation, and it is not appropriate to use such imagery in a public or professional setting.}
    \label{fig:case2}
\end{figure}

Figure~\ref{fig:case2} is a split-image meme. On the top, a middle-aged Black man in a dark suit has a serious expression. On the bottom, a noose made of blue rope is shown against a black background. The caption reads: \textit{"it's what everyone's been waiting for."} 

The model incorrectly classified this image as non-toxic. The likely reason is that the caption did not explicitly mention the man’s identity (Barack Obama), which prevented the model from associating the image with its full context. Without recognizing the specific political or racial implications of the noose imagery, the model misinterpreted the meme's intent. This highlights a challenge in entity recognition—when crucial details are missing from the caption, the model struggles to infer context accurately.

\subsection{Case 3: Confusion Due to Facial Features and Expression}
\label{subsec:case3}
\begin{figure}[h]
    \centering
    \includegraphics[width=0.6\linewidth]{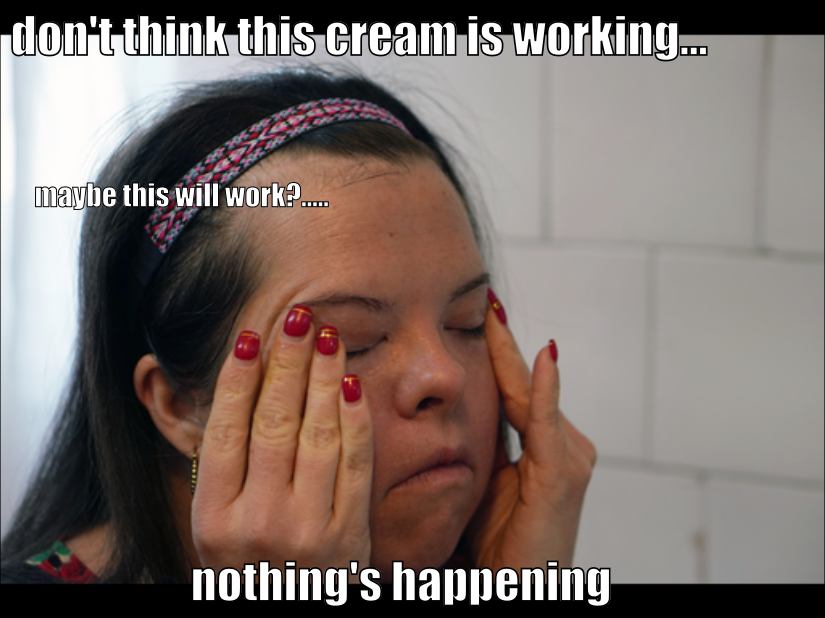}
    \caption{Meme with woman applying cream. \\ \textbf{LLaVA caption:} The meme features a close-up of a person with their eyes closed, hands placed on their face, and a pained or distressed expression. The person appears to be a woman with dark hair, wearing a headband with a patterned design. She has red nail polish on her nails. The background is a plain, light-colored wall. The meme is captioned with a text that reads, \textit{"don't think this cream is working... maybe this will work??... nothing's happening."} The text is in a casual, conversational font, suggesting a humorous or sarcastic tone. The primary subject of the meme seems to be the process of applying or using a product, such as a cream or lotion, with the implication that the product is not working as expected. The person in the image is likely the target of the meme, as they are shown in a state of frustration or disappointment with the product. The meme does not explicitly identify the ethnicity, gender, or other characteristics of the individual, but it is clear that the person is a woman. The meme does not target any specific group or individual based on their ethnicity, gender, or other characteristics. Instead, it focuses on a relatable experience of product use and the potential for disappointment or frustration when a product does not meet expectations. The meme is likely intended to be humorous or relatable to a wide audience, rather than targeting a specific group.}
    \label{fig:case3}
\end{figure}

The meme in Figure~\ref{fig:case3} shows a woman with a headband and red nail polish, pressing her hands against her face with a distressed expression. The caption reads: \textit{"don't think this cream is working... maybe this will work??... nothing's happening."}

The model incorrectly classified this image as toxic. This may have been due to the woman’s facial features—specifically, her relatively thin eyes while being in distress—which led the model to incorrectly associate the image with offensive stereotypes. This suggests that certain facial expressions and features in training data may have been disproportionately linked to harmful content, causing unintended biases in classification. However, the image itself is entirely benign, merely depicting frustration with a skincare product.

\end{document}